\documentclass[nojss]{jss}
\usepackage{thumbpdf,lmodern}
\graphicspath{{Figures/}} 
\usepackage{amsmath}
\usepackage{mathtools}
\usepackage{amsfonts}
\usepackage{graphicx}
\usepackage{caption}
\usepackage{subcaption}
\usepackage{booktabs}
\usepackage{xspace}
\usepackage{subdepth} 
\usepackage{float}
\usepackage{soul} 
\usepackage{enumerate}
\usepackage{array}
\usepackage{multirow}

\newcommand{\norm}[1]{\left\lVert#1\right\rVert}
\newcommand{\goodhat}[1]{\expandafter\hat#1}
\newcommand{\goodtilde}[1]{\expandafter\tilde#1}

\newcommand{\bydef}{\ensuremath{\equiv}\xspace}

\providecommand\bgamma{\mbox{\boldmath $\gamma$}}
\providecommand\bbeta{\mbox{\boldmath $\beta$}}
\providecommand\bx{\mbox{\boldmath $x$}}
\providecommand\bs{\mbox{\boldmath $s$}}

\newcolumntype{L}[1]{>{\raggedright\arraybackslash}p{\dimexpr#1-2\tabcolsep-\arrayrulewidth}}
\newcolumntype{R}[1]{>{\raggedright\arraybackslash}p{\dimexpr#1-2\tabcolsep-\arrayrulewidth}}
\author{David Ardia \\ HEC Montr\'eal \\ GERAD \And
Keven Bluteau \\ Universit\'e de Sherbrooke \AND
Samuel Borms \\ University of Neuch\^atel \\ Vrije Universiteit Brussel \And
Kris Boudt \\ Ghent University \\ Vrije Universiteit Brussel \\ Vrije Universiteit Amsterdam \AND
Journal of Statistical Software, 2021, 99, 1-40\\
https://doi.org/10.18637/jss.v099.i02}
\Plainauthor{D.~Ardia, K.~Bluteau, S.~Borms, K.~Boudt}

\title{The \proglang{R}~Package \pkg{sentometrics} to Compute, Aggregate and
      Predict with Textual Sentiment}
\Plaintitle{The R Package sentometrics to Compute, Aggregate and Predict
          with Textual Sentiment}
\Shorttitle{The \proglang{R}~Package \pkg{sentometrics}}

\Abstract{
We provide a hands-on introduction to optimized textual sentiment
indexation using the \proglang{R}~package \pkg{sentometrics}.  Textual
sentiment analysis is increasingly used to unlock the potential information
value of textual data.  The \pkg{sentometrics} package implements an
intuitive framework to efficiently compute sentiment scores of numerous
texts, to aggregate the scores into multiple time series, and to use these
time series to predict other variables.  The workflow of the package is
illustrated with a built-in corpus of news articles from two major U.S.\
journals to forecast the CBOE Volatility Index.
}

\Keywords{Aggregation, Penalized Regression, Prediction, \proglang{R},
\pkg{sentometrics}, Textual Sentiment, Time Series}
\Plainkeywords{Aggregation, Penalized Regression, R, sentometrics, Textual
Sentiment, Time Series}

\Address{
David Ardia\\
Department of Decision Sciences\\
HEC Montr\'eal, Canada\\
\& GERAD\\
Montr\'eal, Canada\\
E-mail:~\email{david.ardia@hec.ca}\\
\\
Keven Bluteau\\
Department of Finance\\
Universit\'e de Sherbrooke, Canada\\
E-mail:~\email{keven.bluteau@usherbrooke.ca}\\
\\
Samuel Borms (corresponding author) \\
Institute of Financial Analysis\\
University of Neuch\^atel, Switzerland\\
\& Solvay Business School\\
Vrije Universiteit Brussel, Belgium\\
E-mail:~\email{borms\_sam@hotmail.com}\\
\\
Kris Boudt\\
Department of Economics\\
Ghent University, Belgium\\
\& Solvay Business School\\
Vrije Universiteit Brussel, Belgium\\
\& School of Business and Economics\\
Vrije Universiteit Amsterdam, The Netherlands\\
E-mail:~\email{kris.boudt@ugent.be}\\
}

\begin{document}

\section{Introduction}

Individuals, companies, and governments continuously consume written
material from various sources to improve their decisions.  The corpus of
texts is typically of a high-dimensional longitudinal nature requiring
statistical tools to extract the relevant information.  A key source of
information is the sentiment transmitted through texts, called
\emph{textual sentiment}.  \citet{algaba19} review the notion of sentiment
and its applications, mainly in economics and finance.  They define
sentiment as ``the disposition of an entity towards an entity, expressed via
a certain medium.'' The medium in this case is texts.  The sentiment
expressed through texts may provide valuable insights on the future dynamics
of variables related to firms, the economy, political agendas, product
satisfaction, and marketing campaigns, for instance.  Still, textual
sentiment does not live by the premise to be equally useful across all
applications.  Deciphering when, to what degree, and which layers of the
sentiment add value is needed to consistently study the full information
potential present within qualitative communications.  The econometric
approach of constructing time series of sentiment by means of optimized
selection and weighting of textual sentiment is referred to as
\emph{sentometrics} by \citet{algaba19} and \citet{ardia19}.  The term
sentometrics is a composition of (textual) sentiment analysis and (time
series) econometrics.

The release of the \proglang{R} \citep{R} text mining infrastructure
\pkg{tm} \citep{tm} over a decade ago can be considered the starting point
of the development and popularization of textual analysis tools in
\proglang{R}.  A number of successful follow-up attempts at improving the
speed and interface of the comprehensive natural language processing
capabilities provided by \pkg{tm} have been delivered by the packages
\pkg{openNLP} \citep{openNLP}, \pkg{cleanNLP} \citep{cleanNLP},
\pkg{quanteda} \citep{quanteda}, \pkg{tidytext} \citep{tidytext}, and
\pkg{qdap} \citep{qdap}.

The notable tailor-made packages for sentiment analysis in \proglang{R} are
\pkg{meanr} \citep{meanr}, \pkg{SentimentAnalysis}
\citep{sentimentanalysis}, \pkg{sentimentr} \citep{sentimentr}, and
\pkg{syuzhet} \citep{syuzhet}.  Many of these packages rely on one of the
larger above-mentioned textual analysis infrastructures.  The \pkg{meanr}
package computes net sentiment scores fastest, but offers no
flexibility.\footnote{In a supplementary appendix, we provide an
illustrative comparison of the computation time for various lexicon-based
sentiment calculators in \proglang{R}, including the one from the
\pkg{sentometrics} package.  The appendix and the replication script
\code{run\_timings.R} are available on our package's \proglang{GitHub} repository in
the \code{appendix} folder.} The \pkg{SentimentAnalysis} package relies on a
similar calculation as used in \pkg{tm}'s sentiment scoring function.  The
package can additionally be used to generate and evaluate sentiment
dictionaries.  The \pkg{sentimentr} package extends the polarity scoring
function from the \pkg{qdap} package to handle more difficult linguistic
edge cases, but is therefore slower than packages which do not attempt this. 
The \pkg{SentimentAnalysis} and \pkg{syuzhet} packages also become
comparatively slower for large input corpora.  The \pkg{quanteda} and
\pkg{tidytext} packages have no explicit sentiment computation function but
their toolsets can be used to construct one.

Our \proglang{R}~package \pkg{sentometrics} proposes a well-defined
modeling workflow, specifically targeted at studying the evolution of
textual sentiment and its impact on other quantities.  It can be used (i) to
\emph{compute} textual sentiment, (ii) to \emph{aggregate} fine-grained
textual sentiment into various sentiment time series, and (iii) to
\emph{predict} other variables with these sentiment measures.  The
combination of these three facilities leads to a flexible and
computationally efficient framework to exploit the information value of
sentiment in texts.  The package presented in this paper therefore addresses
the present lack of analytical capability to extract time series
intelligence about the sentiment transmitted through a large panel of texts.

Furthermore, the \pkg{sentometrics} package positions itself as both
integrative and supplementary to the powerful text mining and data science
toolboxes in the \proglang{R} universe.  It is integrative, as it combines
the strengths of \pkg{quanteda} and \pkg{stringi} \citep{stringi} for corpus
construction and manipulation.  It uses \pkg{data.table} \citep{dt} for fast
aggregation of textual sentiment into time series, and \pkg{glmnet}
\citep{glmnet} and \pkg{caret} \citep{caret} for (sparse) model estimation. 
It is supplementary, given that it easily extends any text mining workflow
to compute, aggregate and predict with textual sentiment.

The remainder of the paper is structured as follows. 
Section~\ref{s:methodology} introduces the methodology behind the
\proglang{R}~package \pkg{sentometrics}.  Section~\ref{s:package} describes
the main control functions and illustrates the package's typical workflow. 
Section~\ref{s:application} applies the entire framework to forecast the
Chicago Board Options Exchange (CBOE) Volatility Index. 
Section~\ref{s:conclusion} concludes.

\section{Use cases and workflow}\label{s:methodology}

The typical use cases of the \proglang{R}~package \pkg{sentometrics} are the
fast computation and aggregation of textual sentiment, the subsequent time
series visualization and manipulation, and the estimation of a
sentiment-based prediction model.  The use case of building a prediction
model out of textual data encompasses the previous ones.

\begin{table}[t!]
\centering
\begin{tabular}{L{4.75cm}L{7.55cm}L{3.2cm}}
\toprule[1pt]
Functionality & Functions & Output \\
\toprule[1pt]
\textbf{1. Corpus management} &  &  \\
(a) Creation & \code{sento_corpus()} & `\code{sento_corpus}' \\
(b) Manipulation & \pkg{quanteda} corpus functions (e.g.,~\code{docvars()}, \code{corpus_sample()}, or \code{corpus_subset()}), \code{as.data.frame()}, \code{as.data.table()}, \code{as.sento_corpus()} &  \\
(c) Features generation &  \code{add_features()} &  \\
(d) Summarization & \code{corpus_summarize()}, \code{print()} & \\
\addlinespace[1ex]
\multicolumn{2}{l}{\textbf{2. Sentiment computation}} & \\
(a) Lexicon management & \code{sento_lexicons()} & `\code{sento_lexicons}' \\
(b) Computation &  \code{compute_sentiment()} & `\code{sentiment}' \\
(c) Manipulation &  \code{merge()}, \code{as.sentiment()} & \\
(d) Summarization & \code{peakdocs()} & \\
\addlinespace[1ex]
\multicolumn{2}{l}{\textbf{3. Sentiment aggregation}} &  \\
(a) Specification & \code{ctr_agg()} &  \\
(b) Aggregation & \code{sento_measures()}, \code{aggregate()} &
`\code{sento_measures}' \\
(c) Manipulation & \code{subset()}, \code{merge()}, \code{diff()},
\code{scale()}, \code{as.data.frame()}, \code{as.data.table()},
\code{measures_fill()}, \code{measures_update()} &  \\
(d) Visualization & \code{plot()} &  \\
(e) Summarization & \code{summary()}, \code{peakdates()}, \code{print()},
\code{nobs()}, \code{nmeasures()}, \code{get_dimensions()}, \code{get_dates()} &  \\
\addlinespace[1ex]
\textbf{4. Modeling} &  &  \\
(a) Specification & \code{ctr_model()} &  \\
(b) Estimation & \code{sento_model()} & `\code{sento_model}',
`\code{sento_modelIter}'  \\
(c) Prediction & \code{predict()} &  \\
(d) Diagnostics & \code{summary()}, \code{print()}, \code{get_loss_data()},
\code{attributions()} & `\code{attributions}' \\
(e) Visualization & \code{plot()} &  \\
\bottomrule[1pt]      
\end{tabular}
\caption{Taxonomy of the \proglang{R}~package \pkg{sentometrics}.  This
table displays the functionalities of the \pkg{sentometrics} package along
with the associated functions and \proglang{S}3 output objects.  We explain the
(generic) \proglang{R}(-style) methods in Appendix~\ref{appendix:methods}.}
\label{table:taxonomy}
\end{table}
We propose a modular workflow that consists of five main steps,
\emph{Steps 1--5}, from corpus construction to model estimation.  All use
cases can be addressed by following (a subset of) this workflow.  The
\proglang{R}~package \pkg{sentometrics} takes care of all steps, apart from
corpus collection and cleaning.  However, various conversion functions and
method extensions are made available that allow the user to enter and exit
the workflow at different steps.  Table~\ref{table:taxonomy} pieces together
the key functionalities of \pkg{sentometrics} together with the associated
functions and \proglang{S}3 class objects.  All steps are explained below.  We
minimize the mathematical details to clarify the exposition, and stay close
to the actual implementation.  Section~\ref{s:package} demonstrates how to
use the functions.

\subsection{Pre-process a selection of texts and generate relevant features (Step 1)}
We assume the user has a corpus of texts of any size at its disposal.  The
data can be scraped from the web, retrieved from news databases, or obtained
from any other source.  The texts should be cleaned such that graphical and
web-related elements (e.g.,~\proglang{HTML} tags) are removed.  To
benefit from the full functionality of the \pkg{sentometrics} package, a
minimal requirement is that every text has a timestamp and an identifier. 
This results in a set of documents $d_{n, t}$ for $n = 1, \ldots, N_{t}$ and
time points $t = 1, \ldots, T$, where $N_{t}$ is the total number of
documents at time $t$.  If the user has no interest in an aggregation into
time series, desiring to do only sentiment calculation, the identifiers and
especially the timestamps can be dropped.  The corpus can also be given a
language identifier, for a sentiment analysis across multiple languages at
once.  The identifier is used to direct the lexicons in the different
languages to the right texts.

Secondly, \emph{features} have to be defined and mapped to the documents. 
Features can come in many forms: news sources, entities (individuals,
companies or countries discussed in the texts), or text topics.  The mapping
implicitly permits subdividing the corpus into many smaller groups with a
common interest.  Many data providers enrich their textual data with
information that can be used as features.  If this is not the case, topic
modeling or entity recognition techniques are valid alternatives.  Human
classification or manual keyword(s) occurrence searches are simpler options. 
The extraction and inclusion of features is an important part of the
analysis and should be related to the variable that is meant to be
predicted.

The texts and features have to be structured in a rectangular fashion. 
Every row represents a document that is mapped to the features through
numerical values $w_{n, t}^{k} \in [0, 1]$ where the features are indexed by
$k = 1, \ldots, K$.  The values are indicative of the relevance of a feature
to a document.  Binary values indicate which documents belong to which
feature(s).

This rectangular data structure is turned into a
`\code{sento_corpus}' object when passed to the \code{sento_corpus()}
function.  The reason for this separate corpus structure is twofold.  It
controls whether all corpus requirements for further analysis are met
(specifically, dealing with timestamps and numeric features), and it allows
performing operations on the corpus in a more structured way.  If no
features are of interest to the analysis, a dummy feature valued $w_{n,
t}^{k} = 1$ throughout is automatically created.  The \code{add_features()}
function is used to add or generate new features, as will be shown in the
illustration.  When the corpus is constructed, it is up to the user to
decide which texts have to be kept for the actual sentiment analysis.

\subsection{Sentiment computation and aggregation (Steps 2 and 3)}

Overall, in the sentiment computation and aggregation framework, we define
three weighting parameters: $\omega$, $\theta$ and $b$.  They control
respectively the within-document, across-document, and across-time
aggregation.  Section~\ref{s:sento_measures} explains how to set the values
for these parameters.  Appendix~\ref{appendix:weighting} gives a overview of
the implemented formulae for weighting.

\subsubsection{Compute document- or sentence-level textual sentiment (Step
2)}

Every document requires at least one sentiment score for further analysis. 
The \pkg{sentometrics} package can be used to assign sentiment using the
\emph{lexicon}-based approach, possibly augmented with information from
\emph{valence shifters}.  The sentiment computation always starts from a
corpus of documents.  However, the package can also automatically decompose
the documents into sentences and return sentence-level sentiment scores. 
The actual computation of the sentiment follows one of the three approaches
explained below.  Alternatively, one can align own sentiment scores with the
\pkg{sentometrics} package making use of the \code{as.sentiment()} and
\code{merge()} functions.

The lexicon-based approach to sentiment calculation is flexible,
transparent, and computationally convenient.  It looks for words (or
unigrams) that are included in a pre-defined word list of polarized
(positive and negative) words.  The package benefits from built-in word
lists in English, French, and Dutch, with the latter two mostly as a checked
web-based translation.  The \pkg{sentometrics} package allows for three
different ways of doing the lexicon-based sentiment calculation.  These
procedures, though simple at their cores, have proven efficient and powerful
in many applications.  In increasing complexity, the supported approaches
are:
\begin{enumerate}[(i)]
\item A \textbf{unigrams} approach.  The most straightforward method, where
computed sentiment is simply a (weighted) sum of all detected word scores as
they appear in the lexicon.
\item A \textbf{valence-shifting bigrams} approach.  The impact of the word
appearing before the detected word is evaluated as well.  A common example
is ``not good'', which under the default approach would get a score of 1
(``good''), but now ends up, for example, having a score of $-1$ due to the
presence of the negator ``not''.
\item A \textbf{valence-shifting clusters} approach.  Valence shifters can
also appear in positions other than right before a certain word.  We
implement this layer of complexity by searching for valence shifters (and
other sentiment-bearing words) in a cluster of at maximum four words before
and two words after a detected polarized word.
\end{enumerate}
In the first two approaches, the sentiment score of a document $d_{n, t}$
($d$ in short) is the sum of the adjusted sentiment scores of all its
unigrams.  The adjustment comes from applying weights to each unigram based
on its position in the document and adjusting for the presence of a valence
shifting word.  This leads to:
\begin{equation}\label{eq:aggwithin}
s_{n, t}^{\{l\}} \bydef \sum_{i = 1}^{Q_{d}} \omega_{i} v_{i} s_{i, n, t}^{\{l\}},
\end{equation}
for every lexicon $l = 1, \ldots, L$.  The total number of unigrams in the
document is equal to $Q_{d}$.  The score $s_{i, n, t}^{\{l\}}$ is the
sentiment value attached to unigram $i$ from document $d_{n, t}$, based on
lexicon $l$.  It equals zero when the word is not in the lexicon.  The
impact of a valence shifter is represented by $v_{i}$, being the shifting
value of the \emph{preceding} unigram $i-1$.  No valence shifter or the
simple unigrams approach boils down to $v_{i} = 1$.  If the valence shifter
is a negator, typically $v_{i} = -1$.  The weights $\omega_{i}$ define the
within-document aggregation.  The values $\omega_{i}$ and $v_{i}$ are
specific to a document $d_{n, t}$, but we omit the indices $n$ and $t$ for
brevity.

The third approach differs in the way it calculates the impact of valence
shifters.  A document is decomposed into $C_{d}$ clusters around polarized
words, and the total sentiment equals the sum of the sentiment of each
cluster.  The expression~\eqref{eq:aggwithin} becomes in this case $s_{n,
t}^{\{l\}} \bydef \sum_{J = 1}^{C_{d}} s_{J, n, t}^{\{l\}}$.  Given a
detected polarized word, say unigram $j$, valence shifters are identified in
a surrounding cluster of adjacent unigrams $J \bydef \{J^{L}, J^{U}\}$
around this word (irrespective of whether they appear in the same sentence
or not), where $J^{L} \bydef \{j-4, j-3, j-2, j-1\}$ and $J^{U} \bydef
\{j+1, j+2\}$.  The resulting sentiment value of cluster $J$ around
associated unigram $j$ is $s_{J, n, t}^{\{l\}} \bydef n_{N} ( 1 + \max\{0.80
(n_{A} - 2n_{A}n - n_{D}), -1\} ) \omega_{j} s_{j, n, t}^{\{l\}} + \sum_{m
\in J^{U}} \omega_{m} s_{m, n, t}^{\{l\}}$.  The number of amplifying
valence shifters is $n_{A}$, those that deamplify are counted by $n_{D}$, $n
= 1$ and $n_{N} = - 1$ if there is an odd number of negators, else $n = 0$
and $n_{N} = 1$.\footnote{Amplifying valence shifters, such as ``very'',
strengthen a polarized word.  Deamplifying valence shifters, such as
``hardly'', downtone a polarized word.  The strengthening value of 0.80 is
fixed and acts, if applicable, as a modifier of 80\% on the polarized word. 
Negation inverses the polarity.  An even number of negators is assumed to
cancel out the negation.  Amplifiers are considered (but not
double-counted) as deamplifiers if there is an odd number of negators. 
Under this approach, for example, the occurrence of ``not very good'' receives
a more realistic score of $-0.20$ ($n = 1$, $n_{N} = -1$, $n_{A} = 1$,
$n_{D} = 0$, $s = +1$), instead of $-1.80$, in many cases too negative.} All
$n_{A}$, $n_{D}$, $n$ and $n_{N}$ are specific to a cluster $J$.  The
unigrams in $J$ are first searched for in the lexicon, and only when there
is no match, they are searched for in the valence shifters word list. 
Clusters are non-overlapping from one polarized word to the other; if
another polarized word is detected at position $j+4$, then the cluster
consists of the unigrams $\{j+3, j+5, j+6\}$.  This clusters-based approach
borrows from how the \proglang{R}~package \pkg{sentimentr} does its
sentiment calculation.  Linguistic intricacies (e.g.,~sentence
boundaries) are better handled in their package, at the expense of being
slower.

In case of a clusters-based sentence-level sentiment calculation, we
follow the default settings used in \pkg{sentimentr}.  This includes, within
the scope of a sentence, a cluster of 5 words (not 4 as above) before and 2
words after the polarized word, limited to occurring commas.  A fourth type
of valence shifters, adversative conjunctions (e.g.,~however), is
used to reweight the first expression of $\max\{\cdot, -1\}$ by $1 + 0.25
n_{AC}$, where $n_{AC}$ is the difference between the number of adversative
conjunctions within the cluster before and after the polarized word.

The scores obtained above are subsequently multiplied by the feature weights
to spread out the sentiment into lexicon- and feature-specific sentiment
scores, as $s_{n, t}^{\{l, k\}} \bydef s_{n, t}^{\{l\}} w_{n, t}^{k}$, with
$k$ the index denoting the feature.  If the document does not correspond to
the feature, the value of $s_{n, t}^{\{l, k\}}$ is zero.

In \pkg{sentometrics}, the \code{sento_lexicons()} function is used to
define the lexicons and the valence shifters.  The output is a
`\code{sento_lexicons}' object.  Any provided lexicon is applied to
the corpus.  The sentiment calculation is performed with
\code{compute_sentiment()}.  Depending on the input type, this function
outputs a \code{data.table} with all values for $s_{n, t}^{\{l, k\}}$.  When
the output can be used as the basis for aggregation into time series in the
next step (that is, when it has a \code{"date"} column), it becomes a
`\code{sentiment}' object.  To do the computation at sentence-level,
the argument \code{do.sentence = TRUE} should be used.  The
\code{as.sentiment()} function transforms a properly structured table with
sentiment scores into a `\code{sentiment}' object.

\subsubsection{Aggregate the sentiment into textual sentiment time series
(Step 3)}

In this step, the purpose is to aggregate the individual sentiment scores
and obtain various representative time series.  Two main aggregations are
performed.  The first, across-document, collapses all sentiment scores
across documents within the same frequency (e.g.,~day or month, as
defined by $t$) into one score.  The weighted sum that does so is:
\begin{equation}\label{eq:aggdocuments}
s_{t}^{\{l, k\}} \bydef \sum_{n = 1}^{N_{t}} \theta_{n} s_{n, t}^{\{l, k\}}.
\end{equation}
The weights $\theta_{n}$ define the importance of each document $n$ at time
$t$ (for instance, based on the length of the text).  The second,
across-time, smooths the newly aggregated sentiment scores over time, as:
\begin{equation}\label{eq:aggtime}
s_{u}^{\{l, k, b\}} \bydef \sum_{t = t_{\tau}}^{u} b_{t} s_{t}^{\{l, k\}},
\end{equation}
where $t_{\tau} \bydef u - \tau + 1$.  The time weighting schemes $b = 1,
\ldots, B$ go with different values for $b_{t}$ to smooth the time series in
various ways (e.g.,~according to an upward sloping exponential
curve), with a time lag of $\tau$.  The first $\tau - 1$ observations are
dropped from the original time indices, such that $u = \tau, \ldots, T$
becomes the time index for the ultimate time series.  This leaves $N \bydef
T - \tau + 1$ time series observations.

The number of obtained time series, $P$, is equal to $L$ (number of
lexicons) $\times$ $K$ (number of features) $\times$ $B$ (number of time
weighting schemes).  Every time series covers one aspect of each dimension,
best described as ``the textual sentiment as computed by lexicon $l$ for
feature $k$ aggregated across time using scheme $b$''.  The time series are
designated by $s_{u}^{p} \bydef s_{u}^{\{l, k, b\}}$ across all values of
$u$, for $p = 1, \ldots, P$, and the triplet $p \bydef {\{l, k, b\}}$.  The
ensemble of time series captures both different information (different
features) and the same information in different ways (same features,
different lexicons, and aggregation schemes).

The entire aggregation setup is specified by means of the \code{ctr_agg()}
function, including the within-document aggregation needed for the
sentiment analysis.  The \code{sento_measures()} function performs both the
sentiment calculation (via \code{compute_sentiment()}) and time series
aggregation (via \code{aggregate()}), outputting a
`\code{sento_measures}' object.  The obtained sentiment measures in
the `\code{sento_measures}' object can be further aggregated across
measures, also with the \code{aggregate()} function.

\subsection{Specify regression model and do (out-of-sample) predictions
(Step 4)}

The sentiment measures are now regular time series variables that can be
applied in regressions.  In case of a linear regression, the reference
equation is:
\begin{equation}\label{eq:reg}
y_{u + h} = \delta + \bgamma^\top \bx_{u} + \beta_{1}s_{u}^{1} + \ldots + \beta_{p}s_{u}^{p} + \ldots + \beta_{P}s_{u}^{P} + \epsilon_{u + h}.
\end{equation}
The target variable $y_{u + h}$ is often a variable to forecast, that is, $h
> 0$.  Let $\bs_{u} \bydef (s_{u}^{1}, \ldots, s_{u}^{P})^\top$ encapsulate
all textual sentiment variables as constructed before, and $\bbeta \bydef
(\beta_{1}, \ldots, \beta_{P})^\top$.  Other variables are denoted by the
vector $\bx_{u}$ at time $u$ and $\bgamma$ is the associated parameter
vector.  Logistic regression (binomial and multinomial) is available as a
generalization of the same underlying linear structure.

The typical large dimensionality of the number of predictors
in~\eqref{eq:reg} relative to the number of observations, and the potential
multicollinearity, both pose a problem to ordinary least squares (OLS)
regression.  Instead, estimation and variable selection through a penalized
regression relying on the elastic net regularization of \citet{zou05} is
more appropriate.  As an example, \citet{joshi10} and \citet{yogatama11} use
regularization to predict movie revenues, and scientific article downloads
and citations, respectively, using many text elements such as words,
bigrams, and sentiment scores.  \citet{ardia19} similarly forecast U.S.\
industrial production growth based on a large number of sentiment time
series extracted from newspaper articles.

Regularization, in short, shrinks the coefficients of the least informative
variables towards zero.  It consists of optimizing the least squares or
likelihood function including a penalty component.  The elastic net
optimization problem for the specified linear regression is expressed as:
\begin{equation}\label{eq:elnet}
\underset{\tilde{\delta}, \widetilde{\bgamma}, \widetilde{\bbeta}}{\min} \left\{ \frac{1}{N}\sum_{u=\tau}^{T}(y_{u + h} - \tilde{\delta} - \widetilde{\bgamma}^\top \widetilde{\bx}_{u} - \widetilde{\bbeta}^\top \widetilde{\bs}_{u})^2 + \lambda\left[ \alpha\norm{\widetilde{\bbeta}}_{1} + (1-\alpha)\norm{\widetilde{\bbeta}}_{2}^2 \right] \right\}.
\end{equation}
The tilde denotes standardized variables, and $\norm{.}_{p}$ is the
$\ell_{p}$-norm.  The standardization is required for the regularization,
but the coefficients are rescaled back once estimated.  The rescaled
estimates of the model coefficients for the textual sentiment indices are in
$\widehat{\bbeta}$, usually a sparse vector, depending on the severity of
the shrinkage.  The parameter $0 \le \alpha \le 1$ defines the trade-off
between the Ridge \citep{hoerl70}, $\ell_{2}$, and the LASSO
\citep{tibshirani96}, $\ell_{1}$, regularization, respectively for $\alpha =
0$ and $\alpha = 1$.  The $\lambda \ge 0$ parameter defines the level of
regularization.  When $\lambda = 0$, the problem reduces to OLS estimation. 
The two parameters are calibrated in a data-driven way, such that they are
optimal to the regression equation at hand.  The \pkg{sentometrics} package
allows calibration through cross-validation, or based on an information
criteria with the degrees of freedom properly adjusted to the elastic net
context according to \citet{tibshirani12}.

A potential analysis of interest is the sequential estimation of a
regression model and out-of-sample prediction.  For a given sample size $M
< N$, a regression is estimated with $M$ observations and used to predict
some next observation of the target variable.  This procedure is repeated
rolling forward from the first to the last $M$-sized sample, leading to a
series of estimates.  These are compared with the realized values to assess
the (average) out-of-sample prediction performance.

The type of model, the calibration approach, and other modeling decisions
are defined via the \code{ctr_model()} function.  The (iterative) model
estimation and calibration is done with the \code{sento_model()} function
that relies on the \proglang{R}~packages \pkg{glmnet} and \pkg{caret}.  The
user can define here additional (sentiment) values for prediction through
the \code{x} argument.  The output is a `\code{sento_model}' object
(one model) or a `\code{sento_modelIter}' object (a collection of
iteratively estimated `\code{sento_model}' objects and associated
out-of-sample predictions).

A forecaster, however, is not limited to using the models provided through
the \pkg{sentometrics} package; (s)he is free to guide this step to
whichever modeling toolbox available, continuing with the sentiment
variables computed in the previous steps.

\subsection{Evaluate prediction performance and sentiment attributions (Step
5)}

A `\code{sento_modelIter}' object carries an overview of
out-of-sample performance measures relevant to the type of model
estimated.  Plotting the object returns a time series plot comparing the
predicted values with the corresponding observed ones.  A more formal way to
compare the forecasting performance of different models, sentiment-based or
not, is to construct a model confidence set \citep{hansen11}.  This set
isolates the models that are statistically the best regarding predictive
ability, within a confidence level.  To do this analysis, one needs to first
call the function \code{get_loss_data()} which returns a loss data
\code{matrix} from a collection of `\code{sento_modelIter}' objects,
for a chosen loss metric (like squared errors); see \code{?get_loss_data}
for more details.  This loss data \code{matrix} is ready for use by the
\proglang{R}~package \pkg{MCS} \citep{mcs} to create a model confidence set.

The aggregation into textual sentiment time series is entirely linear. 
Based on the estimated coefficients $\widehat{\bbeta}$, every underlying
dimension's sentiment \emph{attribution} to a given prediction can thus be
computed easily.  For example, the attribution of a certain feature $k$ in
the forecast of the target variable at a particular date is the weighted sum
of the model coefficients and the values of the sentiment measures
constructed from $k$.  Attribution can be computed for all features,
lexicons, time-weighting schemes, time lags, and individual documents. 
Through attribution, a prediction is broken down in its respective
components.  The attribution to documents is useful to pick the texts with
the most impact to a prediction at a certain date.  The function
\code{attributions()} computes all types of possible attributions.

\section[The R package sentometrics]{The \proglang{R}~package
\pkg{sentometrics}}\label{s:package}

In what follows, several examples show how to put the steps into practice
using the \pkg{sentometrics} package.  The subsequent sections illustrate
the main workflow, using built-in data, focusing on individual aspects of
it.  Section~\ref{s:corpus} studies corpus management and features
generation.  Section~\ref{s:sentiment} investigates the sentiment
computation.  Section~\ref{s:sento_measures} looks at the aggregation into
time series (including the control function \code{ctr_agg()}). 
Section~\ref{s:manipulation} briefly explains further manipulation of a
`\code{sento_measures}' object.  Section~\ref{s:modeling} regards
the modeling setup (including the control function \code{ctr_model()}) and
attribution.

\subsection[Corpus management and features generation]{Corpus management and
features generation}\label{s:corpus}

The very first step is to load the \proglang{R}~package \pkg{sentometrics}. 
We also load the \pkg{data.table} package as we use it throughout, but
loading it is in general not required.
\begin{CodeChunk}
\begin{CodeInput}
R> library("sentometrics")
R> library("data.table")                                                
\end{CodeInput}
\end{CodeChunk}
We demonstrate the workflow using the \code{usnews} built-in dataset, a
collection of news articles from The Wall Street Journal and The Washington
Post between 1995 and 2014.\footnote{The data originates from
\url{https://appen.com/open-source-datasets}
(previously under ``Economic News Article Tone and Relevance'').} It has a
\code{data.frame} structure and thus satisfies the requirement that the
input texts have to be structured rectangularly, with every row representing
a document.  The data is loaded below.
\begin{CodeChunk}
\begin{CodeInput}
R> data("usnews", package = "sentometrics")
R> class(usnews)                                                
\end{CodeInput}
\begin{CodeOutput}
[1] "data.frame"
\end{CodeOutput}
\end{CodeChunk}
For conversion to a `\code{sento_corpus}' object, the \code{"id"},
\code{"date"}, and \code{"texts"} columns have to come in that order.  One
could also add an optional \code{"language"} column for a multi-language
sentiment analysis (see the multi-language sentiment computation example in
the next section).  All other columns are reserved for features, of type
numeric.  For this particular corpus, there are four original features.  The
first two indicate the news source, the latter two the relevance of every
document to the U.S.\ economy.  The feature values $w_{n, t}^{k}$ are binary
and complementary (when \code{"wsj"} is 1, \code{"wapo"} is 0; similarly for
\code{"economy"} and \code{"noneconomy"}) to subdivide the corpus to create
separate time series.
\begin{CodeChunk}
\begin{CodeInput}
R> head(usnews[, -3])                                                
\end{CodeInput}
\begin{CodeOutput}
         id       date wsj wapo economy noneconomy
1 830981846 1995-01-02   0    1       1          0
2 842617067 1995-01-05   1    0       0          1
3 830982165 1995-01-05   0    1       0          1
4 830982389 1995-01-08   0    1       0          1
5 842615996 1995-01-09   1    0       0          1
6 830982368 1995-01-09   0    1       1          0
\end{CodeOutput}
\end{CodeChunk}
To access the texts, one can simply do \code{usnews[["texts"]]}
(i.e.,~the third column omitted above).  An example of one text is:
\begin{CodeChunk}
\begin{CodeInput}
R> usnews[["texts"]][2029]                                                
\end{CodeInput}
\begin{CodeOutput}
[1] "Dow Jones Newswires NEW YORK -- Mortgage rates rose in the past week
after Fridays employment report reinforced the perception that the economy
is on solid ground, said Freddie Mac in its weekly survey. The average for
30-year fixed mortgage rates for the week ended yesterday, rose to 5.85 
from 5.79 a week earlier and 5.41 a year ago. The average for 15-year 
fixed-rate mortgages this week was 5.38, up from 5.33 a week ago and the 
year-ago 4.69. The rate for five-year Treasury-indexed hybrid adjustable-rate
mortgages, was 5.22, up from the previous weeks average of 5.17. There is no
historical information for last year since Freddie Mac began tracking this 
mortgage rate at the start of 2005."
\end{CodeOutput}
\end{CodeChunk}
The built-corpus is cleaned for non-alphanumeric characters.  To put the
texts and features into a corpus structure, call the \code{sento_corpus()}
function.  If you have no features available, the corpus can still be
created without any feature columns in the input \code{data.frame}, but a
dummy feature called \code{"dummyFeature"} with a score of 1 for all texts
is added to the `\code{sento_corpus}' output object.
\begin{CodeChunk}
\begin{CodeInput}
R> uscorpus <- sento_corpus(usnews)	
R> class(uscorpus)											
\end{CodeInput}
\begin{CodeOutput}
[1] "sento_corpus" "corpus" "character"   
\end{CodeOutput}
\end{CodeChunk}
The \code{sento_corpus()} function creates a `\code{sento_corpus}'
object on top of the \pkg{quanteda}'s package `\code{corpus}' object. 
Hence, many functions from \pkg{quanteda} to manipulate corpora can be
applied to a `\code{sento_corpus}' object as well.  For instance,
\code{quanteda::corpus_subset(uscorpus, date < "2014-01-01")} would limit
the corpus to all articles before 2014.  The presence of the date document
variable (the \code{"date"} column) and all other metadata as numeric
features valued between 0 and 1 are the two distinguishing aspects between a
`\code{sento_corpus}' object and any other corpus-like object in
\proglang{R}.  Having the date column is a requirement for the later
aggregation into time series.  The function \code{as.sento_corpus()}
transforms a \pkg{quanteda} `\code{corpus}' object, a \pkg{tm}
`\code{SimpleCorpus}' object or a \pkg{tm} `\code{VCorpus}'
object into a `\code{sento_corpus}' object; see
\code{?as.sento_corpus} for more details.

To round off \emph{Step 1}, we add two metadata features using the
\code{add_features()} function.  The features \code{uncertainty} and
\code{election} give a score of 1 to documents in which respectively the
word \code{"uncertainty"} or \code{"distrust"} and the specified regular
expression \code{regex} appear.  Regular expressions provide flexibility to
define more complex features, though it can be slow for a large corpus if
too complex.  Overall, this gives $K = 6$ features.  The
\code{add_features()} function is most useful when the corpus starts off
with no additional metadata, i.e.,~the sole feature present is the
automatically created \code{"dummyFeature"} feature.\footnote{To delete a
feature, use \code{quanteda::docvars(corpusVariable, field = "featureName")
<- NULL}.  The \code{docvars()} method is extended for a
`\code{sento\_corpus}' object; for example, if all current features
are deleted, the dummy feature \code{"dummyFeature"} is automatically
added.}
\begin{CodeChunk}
\begin{CodeInput}
R> regex <- paste0("\\bRepublic[s]?\\b|\\bDemocrat[s]?\\b|\\belection\\b|")
R> uscorpus <- add_features(uscorpus,
+    keywords = list(uncertainty = c("uncertainty", "distrust"), 
+        election = regex),
+    do.binary = TRUE, do.regex = c(FALSE, TRUE))
R> tail(quanteda::docvars(uscorpus))
\end{CodeInput}
\begin{CodeOutput}
                date wsj wapo economy noneconomy uncertainty election
842616931 2014-12-22   1    0       1          0           0        0
842613758 2014-12-23   1    0       0          1           0        0
842615135 2014-12-23   1    0       0          1           0        0
842617266 2014-12-24   1    0       1          0           0        0
842614354 2014-12-26   1    0       0          1           0        0
842616130 2014-12-31   1    0       0          1           0        0
\end{CodeOutput}
\end{CodeChunk}
The \code{corpus_summarize()} function is useful to numerically and visually
display the evolution of various parameters within the corpus.
\begin{CodeChunk}
\begin{CodeInput}
R> summ <- corpus_summarize(uscorpus, by = "year")
R> summ$plots$feature_plot + guides(color = guide_legend(nrow = 1))
\end{CodeInput}
\end{CodeChunk}
\begin{figure}[t!]
\centering
\includegraphics[width=\textwidth]{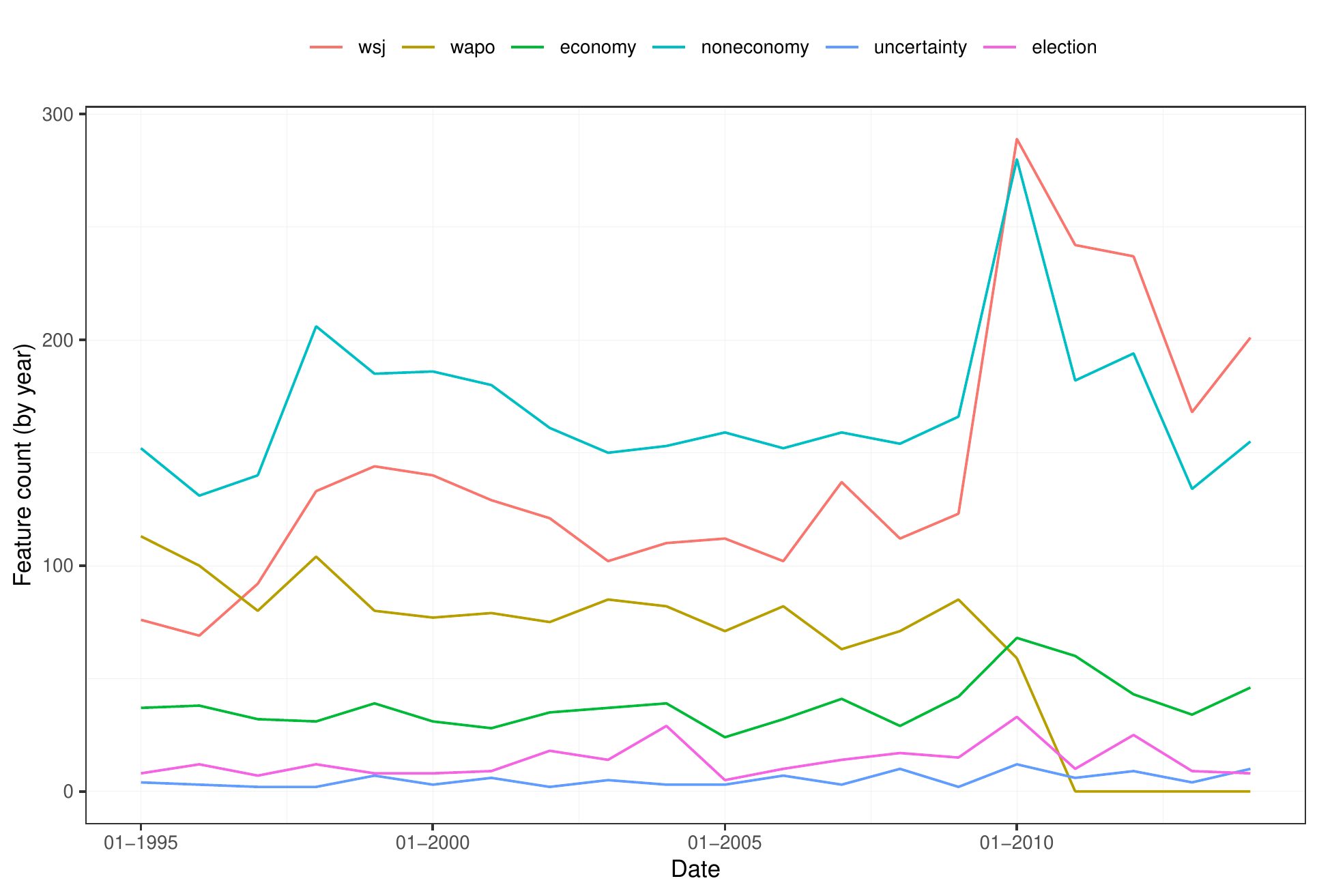}
\caption{Yearly evolution of the features presence across the corpus.}
\label{fig:corpsumm}
\end{figure}
Figure~\ref{fig:corpsumm} shows the counts of the corpus features per
year.\footnote{For more control over the plots, our replication script also
loads the \pkg{ggplot2} \citep{ggplot2} and \pkg{gridExtra}
\citep{gridextra} packages.} The values are obtained by counting a feature
for a document if it is not zero.  Around 2010, the Wall Street Journal and
non-economic news dominated the corpus.

\subsection[Lexicon preparation and sentiment computation]{Lexicon
preparation and sentiment computation}\label{s:sentiment}

As \emph{Step 2}, we calculate sentiment using lexicons.  We supply the
built-in \cite{loughran11} and \cite{henry08}
word lists, and the more generic Harvard General Inquirer word list.  We
also add six other lexicons from the \proglang{R}~package \pkg{lexicon}
\citep{lexicon}: the NRC lexicon \citep{nrc10}, the \cite{hu04} lexicon
, the SentiWord lexicon \citep{sentiword10}, the \cite{syuzhet} lexicon, the SenticNet lexicon \citep{senticnet16}, and the SO-CAL
lexicon \citep{taboada11}.\footnote{You can add any two-column table format
as a lexicon, as long as the first column is of type character and the
second column is numeric.} This gives $L = 9$.

We pack these lexicons together in a named list, and provide it to the
\code{sento_lexicons()} function, together with an English valence word
list.  The \code{valenceIn} argument dictates the complexity of the
sentiment analysis.  If \code{valenceIn = NULL} (the default), sentiment is
computed based on the simplest unigrams approach.  If \code{valenceIn} is a
table with an \code{"x"} and a \code{"y"} column, the valence-shifting
bigrams approach is considered for the sentiment calculation.  The values of
the \code{"y"} column are those used as $v_{i}$.  If the second column is
named \code{"t"}, it is assumed that this column indicates the type of
valence shifter for every word, and thus it forces employing the
valence-shifting clusters approach for the sentiment calculation.  Three
types of valence shifters are supported for the latter method: negators
(value of \code{1}, defines $n$ and $n_{N}$), amplifiers (value of \code{2},
counted in $n_{A}$), and deamplifiers (value of \code{3}, counted in
$n_{D}$).  Adversative conjunctions (value of \code{4}, counted in $n_{AC}$)
are an additional type only picked up during a sentence-level calculation.
\begin{CodeChunk}
\begin{CodeInput}
R> data("list_lexicons", package = "sentometrics")
R> data("list_valence_shifters", package = "sentometrics")
R> lexiconsIn <- c(list_lexicons[c("LM_en", "HENRY_en", "GI_en)],
+    list(NRC = lexicon::hash_sentiment_nrc,
+        HULIU = lexicon::hash_sentiment_huliu,
+        SENTIWORD = lexicon::hash_sentiment_sentiword,
+        JOCKERS = lexicon::hash_sentiment_jockers,
+        SENTICNET = lexicon::hash_sentiment_senticnet,
+        SOCAL = lexicon::hash_sentiment_socal_google))
R> lex <- sento_lexicons(lexiconsIn = lexiconsIn,
+    valenceIn = list_valence_shifters[["en"]])
\end{CodeInput}
\end{CodeChunk}
The \code{lex} output is a `\code{sento_lexicons}' object, which is a
list.  Duplicates are removed, all words are put to lowercase and only
unigrams are kept.
\begin{CodeChunk}
\begin{CodeInput}
R> lex[["HENRY_en"]]
\end{CodeInput}
\begin{CodeOutput}
                 x     y
  1:         above     1
  2:    accomplish     1
  3:  accomplished     1
  4:  accomplishes     1
  5: accomplishing     1
 ---                    
185:         worse    -1
186:        worsen    -1
187:     worsening    -1
188:       worsens    -1
189:         worst    -1
\end{CodeOutput}
\end{CodeChunk}
The individual word lists themselves are \code{data.table}s, as displayed
above.

\subsubsection*{Document-level sentiment computation}

The simplest way forward is to compute sentiment scores for every text in
the corpus.  This is handled by the \code{compute_sentiment()} function,
which works with either a character vector, a `\code{sento_corpus}'
object, a \pkg{quanteda} `\code{corpus}' object, a \pkg{tm}
`\code{SimpleCorpus}' object, or a \pkg{tm} `\code{VCorpus}'
object.  The core of the sentiment computation is implemented in
\proglang{C++} through \pkg{Rcpp} \citep{rcpp}.  The
\code{compute_sentiment()} function has, besides the input corpus and the
lexicons, other arguments.  The main one is the \code{how} argument, to
specify the within-document aggregation.  In the example below, \code{how =
"proportional"} divides the net sentiment score by the total number of
tokenized words.  More details on the contents of these arguments are
provided in Section~\ref{s:sento_measures}, when the \code{ctr_agg()}
function is discussed.  See below for a brief usage and output example.
\begin{CodeChunk}
\begin{CodeInput}
R> sentScores <- compute_sentiment(usnews[["texts"]], 
+    lexicons = lex, how = "proportional")
R> head(sentScores[, c("id", "word_count", "GI_en", "SENTIWORD", "SOCAL")])
\end{CodeInput}
\begin{CodeOutput}
      id word_count     GI_en SENTIWORD   SOCAL
1: text1        213 -0.013146 -0.003436 0.20679
2: text2        110  0.054545  0.004773 0.15805
3: text3        202  0.004950 -0.006559 0.11298
4: text4        153  0.006536  0.021561 0.15393
5: text5        245  0.008163 -0.017786 0.08997
6: text6        212  0.014151 -0.009316 0.03506
\end{CodeOutput}
\end{CodeChunk}
For a character vector input, the \code{compute_sentiment()} function
returns a \code{data.table} with an identifier column, a word count column
and the computed sentiment scores for all lexicons, as no features are
involved.  When the input is a `\code{sento_corpus}' object, the
output is a `\code{sentiment}' object.  The \pkg{tm}
`\code{SimpleCorpus}' and `\code{VCorpus}' objects are treated
as a character vector input.  Both could be transformed into a
`\code{sento_corpus}' object with the \code{as.sento_corpus()}
function.  A \pkg{tm} `\code{VCorpus}' object as input, for instance,
thus leads to a plain \code{data.table} object, with a similar structure as
above.
\begin{CodeChunk}
\begin{CodeInput}
R> reuters <- system.file("texts", "crude", package = "tm")
R> tmVCorp <- tm::VCorpus(tm::DirSource(reuters),
+    list(reader = tm::readReut21578XMLasPlain))
R> class(compute_sentiment(tmVCorp, lex))
\end{CodeInput}
\begin{CodeOutput}
[1] "data.table" "data.frame"
\end{CodeOutput}
\end{CodeChunk}
In the example below, we exhibit the use of the \code{as.sento_corpus()} and
\code{as.sentiment()} functions, showing how to integrate computations from
another package into the workflow.
\begin{CodeChunk}
\begin{CodeInput}
R> sentoSent <- compute_sentiment(as.sento_corpus(tmVCorp, 
+    dates = as.Date("1993-03-06") + 1:20), lex, "UShaped")
R> tmSentPos <- sapply(tmVCorp, tm::tm_term_score, lex$NRC[y > 0, x])
R> tmSentNeg <- sapply(tmVCorp, tm::tm_term_score, lex$NRC[y < 0, x])
R> tmSent <- cbind(sentoSent[, 1:3], "tm_NRC" = tmSentPos - tmSentNeg)
R> sent <- merge(sentoSent, as.sentiment(tmSent))
R> sent[6:9, c(1, 11:13)]
\end{CodeInput}
\begin{CodeOutput}
    id SENTICNET--dummyFeature SOCAL--dummyFeature tm_NRC--dummyFeature
1: 236                  0.1611              0.3927                   -2
2: 237                  0.0542              0.3441                   10
3: 242                  0.0025              0.1547                   -3
4: 246                  0.2203              0.2581                   -4
\end{CodeOutput}
\end{CodeChunk}
The \code{tmSent} table is appropriately formatted for conversion into a
`\code{sentiment}' object.  The \code{sent} object embeds the
different scoring approaches and can be used for aggregation into time
series following the remainder of the \pkg{sentometrics} workflow.

\subsubsection*{Sentence-level sentiment computation}

A sentiment calculation at sentence-level instead of the given corpus unit
level requires to set \code{do.sentence = TRUE} in the
\code{compute_sentiment()} function.
\begin{CodeChunk}
\begin{CodeInput}
R> sSentences <- compute_sentiment(uscorpus, lex, do.sentence = TRUE)
R> sSentences[1:11, 1:6]
\end{CodeInput}
\begin{CodeOutput}
           id sentence_id       date word_count LM_en--wsj LM_en--wapo 
 1: 830981846           1 1995-01-02         12          0     0.08333        
 2: 830981846           2 1995-01-02         17          0     0.00000        
 3: 830981846           3 1995-01-02         13          0     0.00000        
 4: 830981846           4 1995-01-02         19          0    -0.05263       
 5: 830981846           5 1995-01-02         19          0    -0.15789       
 6: 830981846           6 1995-01-02         40          0     0.00500        
 7: 830981846           7 1995-01-02         49          0    -0.01633       
 8: 830981846           8 1995-01-02         30          0     0.00000        
 9: 830981846           9 1995-01-02         21          0     0.00000        
10: 830981846          10 1995-01-02          8          0     0.00000        
11: 842617067           1 1995-01-05         19          0     0.00000        
\end{CodeOutput}
\end{CodeChunk}
The obtained sentence-level `\code{sentiment}' object can be
aggregated into document-level scores using the \code{aggregate()} function
with \code{do.full = FALSE}.  The value \code{do.full = TRUE} (by default)
aggregates the scores into sentiment time series.  The aggregation across
sentences from the same document is set through the \code{howDocs}
aggregation parameter.
\begin{CodeChunk}
\begin{CodeInput}
R> aggDocuments <- aggregate(sSentences, ctr_agg(howDocs = "equal_weight"), 
+    do.full = FALSE)
R> aggDocuments[1:2, 1:6]
\end{CodeInput}
\begin{CodeOutput}
          id       date word_count LM_en--wsj LM_en--wapo LM_en--economy 
1: 830981846 1995-01-02        228    0.00000     -0.0277        -0.0277           
2: 842617067 1995-01-05        122    0.06471      0.0000         0.0000           
\end{CodeOutput}
\end{CodeChunk}
The example shows the aggregated sentiment scores for document with
identifier \code{"830981846"} as a simple average of the sentiment scores of
its ten sentences.

\subsubsection*{Multi-language sentiment computation}

To run the sentiment analysis for multiple languages, the
`\code{sento_corpus}' object needs to have a character
\code{"language"} identifier column.  The names should map to a named list
of `\code{sento_lexicons}' objects to be applied to the different
texts.  The language information should be expressed in the different unique
lexicon names.  The values for the columns pertaining to a lexicon in
another language than the document are set to zero.
\begin{CodeChunk}
\begin{CodeInput}
R> usnewsLang <- usnews[1:5, 1:3]
R> usnewsLang[["language"]] <- c("fr", "en", "en", "fr", "en")
R> corpusLang <- sento_corpus(corpusdf = usnewsLang)
R> sLang <- compute_sentiment(corpusLang, list(
+    en = sento_lexicons(list("GI_en" = list_lexicons$GI_en)), 
+    fr = sento_lexicons(list("GI_fr" = list_lexicons$GI_fr_tr))))
R> head(sLang)
\end{CodeInput}
\begin{CodeOutput}
          id       date word_count GI_fr--dummyFeature GI_en--dummyFeature
1: 830981846 1995-01-02        213            0.004695            0.000000
2: 842617067 1995-01-05        110            0.000000            0.054545
3: 830982165 1995-01-05        202            0.000000            0.004950
4: 830982389 1995-01-08        153            0.000000            0.000000
5: 842615996 1995-01-09        245            0.000000            0.008163
\end{CodeOutput}
\end{CodeChunk}

\subsection[Creation of sentiment measures]{Creation of sentiment
measures}\label{s:sento_measures}

\begin{figure}[t!]
\centering
\includegraphics[width=\textwidth]{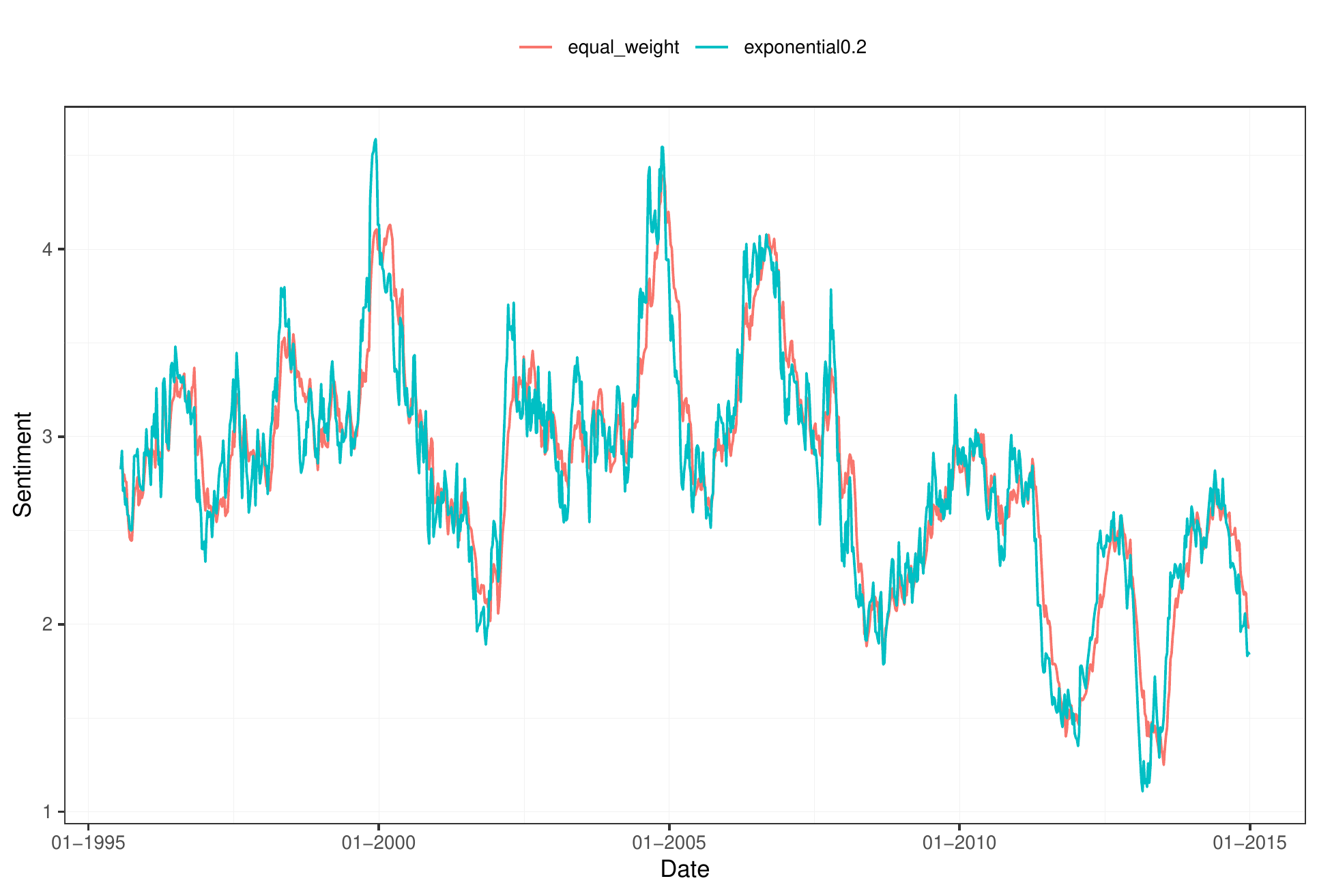}
\caption{Textual sentiment time series averaged across time weighting schemes.}
\label{fig:sentmeasT}
\end{figure}

To create sentiment time series, one needs a well-specified aggregation
setup defined via the control function \code{ctr_agg()}.  To compute the
measures in one go, the \code{sento_measures()} function is to be used. 
Sentiment time series allow to use the entire scope of the package.  We
focus the explanation on the control function's central arguments and
options, and integrate the other arguments in their discussion:

\begin{itemize}
\item[$\bullet$] \code{howWithin}: This argument defines how sentiment is
aggregated within the same document (or sentence), setting the weights
$\omega_{i}$ in~\eqref{eq:aggwithin}.  It is passed on to the \code{how}
argument of the \code{compute_sentiment()} function.  For binary lexicons
and the simple unigrams matching case, the \code{"counts"} option gives
sentiment scores as the difference between the number of positive and
negative words.  Two common normalization schemes are dividing the sentiment
score by the total number of words (\code{"proportional"}) or by the number
of polarized words (\code{"proportionalPol"}) in the document (or sentence). 
A wide number of other weighting schemes are available.  They are, together
with those for the next two arguments, summarized in
Appendix~\ref{appendix:weighting}.
\item[$\bullet$] \code{howDocs}: This argument defines how sentiment is
aggregated across all documents at the same date (or frequency), that is, it
sets the weights $\theta_{n}$ in \eqref{eq:aggdocuments}.  The time
frequency at which the time series have to be aggregated is chosen via the
\code{by} argument, and can be set to daily (\code{"day"}), weekly
(\code{"week"}), monthly (\code{"month"}) or yearly (\code{"year"}).  The
option \code{"equal_weight"} gives the same weight to every document, while
the option \code{"proportional"} gives higher weights to documents with more
words, relative to the document population at a given date.  The
\code{do.ignoreZeros} argument forces ignoring documents with zero sentiment
in the computation of the across-document weights.  By default these
documents are overlooked.  This avoids the incorporation of documents not
relevant to a particular feature (as in those cases $s_{n, t}^{\{l, k\}}$ is
exactly zero, because $w_{n, t}^{k} = 0$), which could lead to a bias of
sentiment towards zero.\footnote{It also ignores the documents which are
relevant to a feature, but exhibit zero sentiment.  This can occur if none
of the words have a polarity, or the weighted number of positive and
negative words offset each other.} When applicable, this argument also
defines the aggregation across sentences within the same document.
\item[$\bullet$] \code{howTime}: This argument defines how sentiment is
aggregated across dates, to smoothen the time series and to acknowledge that
sentiment at a given point is at least partly based on sentiment and
information from the past.  The \code{lag} argument has the role of $\tau$
dictating how far to go back.  In the implementation, \code{lag = 1} means
no time aggregation and thus $b_{t} = 1$.  The \code{"equal_weight"} option
is similar to a simple weighted moving average, \code{"linear"} and
\code{"exponential"} are two options which give weights to the observations
according to a linear or an exponential curve, \code{"almon"} does so based
on Almon polynomials, and \code{"beta"} based on the Beta weighting curve
from \citet{ghysels07}.  The last three curves have respective arguments to
define their shape(s), being \code{alphasExp} and \code{do.inverseExp},
\code{ordersAlm} and \code{do.inverseAlm}, and \code{aBeta} and
\code{bBeta}.  These weighting schemes are always normalized to unity.  If
desired, user-constructed weights can be supplied via \code{weights} as a
named \code{data.frame}.  All the weighting schemes define the different
$b_{t}$ values in \eqref{eq:aggtime}.  The \code{fill} argument is of
sizeable importance here.  It is used to add in dates for which not a single
document was available.  These added, originally missing, dates are given a
value of 0 (\code{"zero"}) or the most recent value (\code{"latest"}).  The
option \code{"none"} accords to not filling up the date sequence at all. 
Adding in dates (or not) impacts the time aggregation by respectively
combining the latest \emph{consecutive} dates, or the latest
\emph{available} dates.
\item[$\bullet$] \code{nCore}: The \code{nCore} argument can help to speed
up the sentiment calculation when dealing with a large corpus.  It expects a
positive integer passed on to the \code{setThreadOptions()} function from
the \pkg{RcppParallel} package \citep{rcppparallel}, and parallelizes the
sentiment computation across texts.  By default, \code{nCore = 1}, which
indicates no parallelization.  Parallelization is expected to improve the
speed of the sentiment computation only for sufficiently large corpora, or
when using many lexicons.
\item[$\bullet$] \code{tokens}: Our unigram tokenization is done with the
\proglang{R}~package \pkg{stringi}; it transforms all tokens to lowercase,
strips punctuation marks and strips numeric characters (see the internal
function \code{sentometrics:::tokenize_texts()}).  If wanted, the texts
could be tokenized separately from the \pkg{sentometrics} package, using any
desired tokenization setup, and then passed to the \code{tokens} argument. 
This way, the tokenization can be tailor-made (\emph{e.g.},
stemmed\footnote{In this case, also stem the lexical entries before you
provide these to the \code{sento\_lexicons()} function.}) and reused for
different sentiment computation function calls, for example to compare the
impact of several normalization or aggregation choices for the same
tokenized corpus.  Doing the tokenization once for multiple subsequent
computation calls is more efficient.  In case of a document-level
calculation, the input should be a list of unigrams per document.  If at
sentence-level (\code{do.sentence = TRUE}), it should be a list of
tokenized sentences as a list of the respective tokenized unigrams.
\end{itemize}
In the example code below, we aggregate sentiment at a weekly frequency,
choose a counts-based within-document aggregation, and weight the
documents for across-document aggregation proportionally to the number of
words in the document.  The resulting time series are smoothed according to
an equally-weighted and an exponential time aggregation scheme ($B = 2$),
using a lag of 30 weeks.  We ignore documents with zero sentiment for
across-document aggregation, and fill missing dates with zero before the
across-time aggregation, as per default.
\begin{CodeChunk}
\begin{CodeInput}
R> ctrAgg <- ctr_agg(howWithin = "counts", howDocs = "proportional",
+    howTime = c("exponential", "equal_weight"), do.ignoreZeros = TRUE, 
+    by = "week", fill = "zero", lag = 30, alphasExp = 0.2)
\end{CodeInput}
\end{CodeChunk}
The \code{sento_measures()} function performs both the sentiment calculation
in \emph{Step 2} and the aggregation in \emph{Step 3}, and results in a
`\code{sento_measures}' output object.  The generic \code{summary()}
displays a brief overview of the composition of the sentiment time series. 
A `\code{sento_measures}' object is a list with as most important
elements \code{"measures"} (the textual sentiment time series),
\code{"sentiment"} (the original sentiment scores per document) and
\code{"stats"} (a selection of summary statistics).  Alternatively, the same
output can be obtained by applying the \code{aggregate()} function on the
output of the \code{compute_sentiment()} function, if the latter is computed
from a `\code{sento_corpus}' object.
\begin{CodeChunk}
\begin{CodeInput}
R> sentMeas <- sento_measures(uscorpus, lexicons = lex, ctr = ctrAgg)
\end{CodeInput}
\end{CodeChunk}
There are 108 initial sentiment measures (9 lexicons $\times$ 6 features
$\times$ 2 time weighting schemes).  An example of one created sentiment
measure and its naming (each dimension's component is separated by
\code{"{-}{-}{-}"}), is shown below.
\begin{CodeChunk}
\begin{CodeInput}
R> as.data.table(sentMeas)[, c(1, 23)]
\end{CodeInput}
\begin{CodeOutput}
            date NRC--noneconomy--equal_weight
   1: 1995-07-24                         3.916
   2: 1995-07-31                         4.109
   3: 1995-08-07                         3.945
   4: 1995-08-14                         4.044
   5: 1995-08-21                         3.831
  ---                                         
1011: 2014-12-01                         4.694
1012: 2014-12-08                         4.694
1013: 2014-12-15                         4.639
1014: 2014-12-22                         4.698
1015: 2014-12-29                         4.876                             
\end{CodeOutput}
\end{CodeChunk}
A `\code{sento_measures}' object is easily plotted across each of its
dimensions.  For example, Figure~\ref{fig:sentmeasT} shows a time series of
average sentiment for both time weighting schemes involved.\footnote{The
averaged sentiment measures can be accessed from the plot object.  Given
\code{p <- plot(\dots)}, this is via \code{p[["data"]]}.  The data is in long
format.} A display of averages across lexicons and features is achieved by
altering the \code{group} argument from the \code{plot()} method to
\code{"lexicons"} and \code{"features"}, respectively.
\begin{CodeChunk}
\begin{CodeInput}
R> plot(sentMeas, group = "time")
\end{CodeInput}
\end{CodeChunk}

\subsection[Manipulation of a sento_measures object]{Manipulation of a
`\code{sento\_measures}' object}\label{s:manipulation}

There are a number of methods and functions implemented to facilitate the
manipulation of a `\code{sento_measures}' object.  Useful methods are
\code{subset()}, \code{diff()}, and \code{scale()}.  The \code{subset()}
function can be used to subset rows by index or by a condition, as well as
to select or delete certain sentiment measures.  The \code{scale()} function
can take \code{matrix} inputs, handy to add, deduct, multiply or divide the
sentiment measures up to the level of single observations.  The functions
\code{get_dates()} and \code{get_dimensions()} are convenient functions that
return the dates and the aggregation dimensions of a
`\code{sento_measures}' object.  The functions \code{nobs()} and
\code{nmeasures()} give the number of time series observations and the
number of sentiment measures.  The \code{as.data.table()} function extracts
the sentiment measures, either in long or wide format.

\begin{figure}[t!]
\centering
\includegraphics[width=\textwidth]{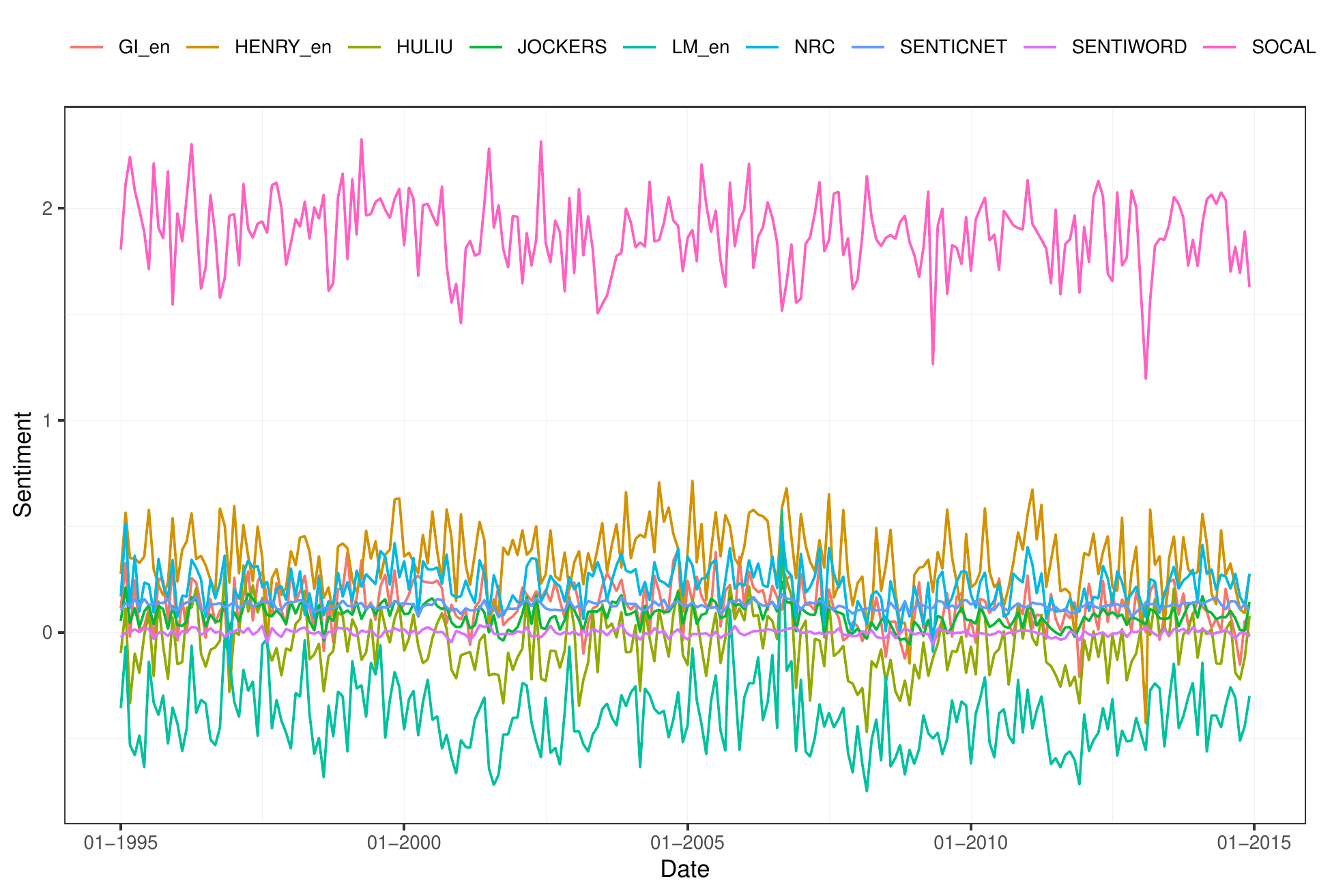}
\caption{Textual sentiment time series averaged across lexicons.}
\label{fig:sentmeasL}
\end{figure}

The \code{select} and \code{delete} arguments in the \code{subset()}
function indicate which combinations of sentiment measures to extract or
delete.  Here, the \code{subset()} function call returns a new object
without all sentiment measures created from the \code{"LM_en"} lexicon and
without the single time series consisting of the specified combination
\code{"SENTICNET"} (lexicon), \code{"economy"} (feature), and
\code{"equal_weight"} (time weighting).\footnote{The new number of sentiment
measures is not necessarily equal to $L \times K \times B$ anymore once a
`\code{sento\_measures}' object is modified.}
\begin{CodeChunk}
\begin{CodeInput}
R> subset(sentMeas, 1:600, delete = list(c("LM_en"),
+    c("SENTICNET", "economy", "equal_weight")))
\end{CodeInput}
\begin{CodeOutput}
A sento_measures object (95 textual sentiment time series, 600 observations).
\end{CodeOutput}
\end{CodeChunk}
Subsetting across rows is done with the \code{subset()} function without
specifying an exact argument.  A typical example is to subset only a time
series range by specifying the dates part of that range, as below, where 50
dates are kept.  One can also condition on specific sentiment measures being
above, below, or between certain values, or directly indicate the row
indices (as shown above).
\begin{CodeChunk}
\begin{CodeInput}
R> subset(sentMeas, date %in% get_dates(sentMeas)[50:99])
\end{CodeInput}
\begin{CodeOutput}
A sento_measures object (108 textual sentiment time series, 50 observations).    
\end{CodeOutput}
\end{CodeChunk}
To \emph{ex-post} fill the time series date-wise, the
\code{measures_fill()} can be used.  The function at minimum fills in
missing dates between the existing date range at the prevailing frequency. 
Dates before the earliest and after the most recent date can be added too. 
The argument \code{fill = "zero"} sets all added dates to zero, whereas
\code{fill = "latest"} takes the most recent known value.  This function is
applied internally depending on the \code{fill} parameter from the
\code{ctr_agg()} function.  The example below pads the time series with
trailing dates, taking the first value that occurs.
\begin{CodeChunk}
\begin{CodeInput}
R> sentMeasFill <- measures_fill(sentMeas, fill = "latest", 
+    dateBefore = "1995-07-01")
R> head(as.data.table(sentMeasFill)[, 1:3])
\end{CodeInput}
\begin{CodeOutput}
         date LM_en--wsj--equal_weight LM_en--wapo--equal_weight
1: 1995-06-26                   -2.021                    -3.064
2: 1995-07-03                   -2.021                    -3.064
3: 1995-07-10                   -2.021                    -3.064
4: 1995-07-17                   -2.021                    -3.064
5: 1995-07-24                   -2.021                    -3.064
6: 1995-07-31                   -2.061                    -3.059
\end{CodeOutput}
\end{CodeChunk}
The sentiment visualized using the \code{plot()} function when there are
many different lexicons, features, and time weighting schemes may give a
distorted image due to the averaging.  To obtain a more nuanced picture of
the differences in one particular dimension, one can ignore the other two
dimensions.  For example, \code{corpusPlain} below has only the dummy
feature, and there is no time aggregation involved (\code{lag = 1}).  This
leaves the lexicons as the sole distinguishing dimension between the
sentiment time series.
\begin{CodeChunk}
\begin{CodeInput}
R> corpusPlain <- sento_corpus(usnews[, 1:3])
R> ctrAggLex <- ctr_agg(howWithin = "proportionalPol", howTime = "own",
+    howDocs = "equal_weight", by = "month", fill = "none", lag = 1)
R> sentMeasLex <- sento_measures(corpusPlain, 
+    lexicons = lex[-length(lex)], ctr = ctrAggLex)
R> mean(as.numeric(sentMeasLex$stats["meanCorr", ]))
\end{CodeInput}
\begin{CodeOutput}
[1] 0.3598
\end{CodeOutput}
\end{CodeChunk}
Figure~\ref{fig:sentmeasL} plots the nine time series, each belonging to a
lexicon.  There are differences in levels.  For example, the \cite{loughran11} lexicon's time series lies almost exclusively in the negative
sentiment area.  The overall average correlation is close to 36\%.

One can do additional aggregation of the sentiment time series with, again,
the \code{aggregate()} function.  This functionality equips the user with a
way to either further diminish or further expand the dimensionality of the
sentiment measures.  For instance, all sentiment measures composed of three
particular time aggregation schemes can be shrunk together, by averaging
across each fixed combination of the other two dimensions, resulting in a
set of new measures.

In the example, both built-in lexicons, both news sources, both added
features, and similar time weighting schemes are collapsed into their
respective new counterparts.  The \code{do.keep = FALSE} option indicates
that the original measures are not kept after merging, such that the number
of sentiment time series effectively goes down to $7 \times 4 \times 1 =
28$.
\begin{CodeChunk}
\begin{CodeInput}
R> sentMeasAgg <- aggregate(sentMeas,
+    time = list(W = c("equal_weight", "exponential0.2")),
+    lexicons = list(LEX = c("LM_en", "HENRY_en", "GI_en")),
+    features = list(JOUR = c("wsj", "wapo"), 
+        NEW = c("uncertainty", "election")), 
+    do.keep = FALSE)
R> get_dimensions(sentMeasAgg)
\end{CodeInput}
\begin{CodeOutput}
$features
[1] "economy" "noneconomy" "JOUR" "NEW"

$lexicons
[1] "NRC" "HULIU" "SENTIWORD" "JOCKERS" "SENTICNET" "SOCAL" "LEX"      

$time
[1] "W"   
\end{CodeOutput}
\end{CodeChunk}
The \code{aggregate()} function with argument \code{do.global = TRUE} merges
the sentiment measures into single dimension-specific time series we refer
to as global sentiment.  Each of the different components of the dimensions
has to receive a weight that stipulates the importance and sign in the
global sentiment index.  The weights need not sum to one, that is, no
condition is imposed.  For example, the global lexicon-oriented sentiment
measure is computed as $s_{u}^{G, L} \bydef \frac{1}{P}\sum_{p =
1}^{P}s_{u}^{p} w_{l, p}$.  The weight to lexicon $l$ that appears in $p$ is
denoted by $w_{l, p}$.  An additional, ``fully global'', measure is composed
as $s_{u}^{G} \bydef (s_{u}^{G, L} + s_{u}^{G, K} + s_{u}^{G, B})/3$.

In the code excerpt, we define \emph{ad-hoc} weights that emphasize the
initial features and the Henry lexicon.  Both time weighting schemes are
weighted equally.\footnote{This can also be achieved by setting \code{time =
1} (similarly for the other arguments), and is in fact the default.} The
output is a four-column \code{data.table}.  The global sentiment measures
provide a low-dimensional summary of the sentiment in the corpus, depending
on some preset parameters related to the importance of the dimensions.
\begin{CodeChunk}
\begin{CodeInput}
R> glob <- aggregate(sentMeas,
+    lexicons = c(0.10, 0.40, 0.05, 0.05, 0.08, 0.08, 0.08, 0.08, 0.08),
+    features = c(0.20, 0.20, 0.20, 0.20, 0.10, 0.10),
+    time = c(1/2, 1/2), do.global = TRUE)
\end{CodeInput}
\end{CodeChunk}
The \code{peakdates()} function pinpoints the dates with most extreme
sentiment (negatively, positively, or both).  The example below extracts the
date with the lowest sentiment time series value across all measures.  The
\code{peakdocs()} function can be applied equivalently to a
`\code{sentiment}' object to get the document identifiers with most
extreme document-level sentiment.
\begin{CodeChunk}
\begin{CodeInput}
R> peakdates(sentMeas, n = 1, type = "neg")
\end{CodeInput}
\begin{CodeOutput}
[1] "2008-01-28"
\end{CodeOutput}
\end{CodeChunk}

\subsection[Sparse regression using the sentiment measures]{Sparse
regression using the sentiment measures}\label{s:modeling}

\emph{Step 4} consists of the regression modeling.  The
\pkg{sentometrics} package offers an adapted interface to sparse regression. 
Other model frameworks can be explored with as input the sentiment measures
extracted through the \code{as.data.table()} (or \code{as.data.frame()})
function.  For example, one could transform the computed sentiment time
series into a `\code{zoo}' object from the \pkg{zoo} package
\citep{zoo}, use any of \pkg{zoo}'s functionalities thereafter (\emph{e.g.},
dealing with an irregular time series structure), or run a simple linear
regression (here on the first six sentiment variables) as follows:
\begin{CodeChunk}
\begin{CodeInput}
R> y <- rnorm(nobs(sentMeas))
R> dt <- as.data.table(sentMeas)
R> z <- zoo::zoo(dt[, !"date"], order.by = dt[["date"]])
R> reg <- lm(y ~ z[, 1:6])
\end{CodeInput}
\end{CodeChunk}
We proceed by explaining the available modeling setup in the
\pkg{sentometrics} package.  The \code{ctr_model()} function defines the
modeling setup.  The main arguments are itemized, all others are reviewed
within the discussion:

\begin{itemize}
\item[$\bullet$] \code{model}: The \code{model} argument can take
\code{"gaussian"} (for linear regression), and \code{"binomial"} or
\code{"multinomial"} (both for logistic regression).  The argument
\code{do.intercept = TRUE} fits an intercept.
\item[$\bullet$] \code{type}: The \code{type} specifies the calibration
procedure to find the most appropriate $\alpha$ and $\lambda$ in
\eqref{eq:elnet}.  The options are \code{"cv"} (cross-validation) or one of
three information criteria (\code{"BIC"}, \code{"AIC"} or \code{"Cp"}).  The
information criterion approach is available in case of a linear regression
only.  The argument \code{alphas} can be altered to change the possible
values for \code{alpha}, and similarly so for the \code{lambdas} argument. 
If \code{lambdas = NULL}, the possible values for $\lambda$ are generated
internally by the \code{glmnet()} function from the \proglang{R}~package
\pkg{glmnet}.  If \code{lambdas = 0}, the regression procedure is OLS.  The
arguments \code{trainWindow}, \code{testWindow} and \code{oos} are needed
when model calibration is performed through cross-validation, that is, when
\code{type = "cv"}.  The cross-validation implemented is based on the
``rolling forecasting origin'' principle, considering we are working with
time series.\footnote{As an example, take 120 observations in total,
\code{trainWindow = 80}, \code{testWindow = 10} and \code{oos = 5}.  In the
first round of cross-validation, a model is estimated for a certain
$\alpha$ and $\lambda$ combination with the first 80 observations, then 5
observations are skipped, and predictions are generated for observations 86
to 95.  The next round does the same but with all observations moved one
step forward.  This is done until the end of the total sample is reached,
and repeated for all possible parameter combinations, relying on the
\code{train()} function from the \proglang{R}~package \pkg{caret}.  The
optimal $(\alpha, \lambda)$ couple is the one that induces the lowest
average prediction error (measured by the root mean squared error for linear
models, and overall accuracy for logistic models).} The argument
\code{do.progress = TRUE} prints calibration progress statements.  The
\code{do.shrinkage.x} argument is a logical vector to indicate on which
external explanatory variables to impose regularization.  These variables,
$x_{u}$, are added through the \code{x} argument of the \code{sento_model()}
function.
\item[$\bullet$] \code{h}: The integer argument \code{h} shifts the response
variable up to $y_{u + h}$ and aligns the explanatory variables in
accordance with \eqref{eq:reg}.\footnote{If the input response variable is
not aligned time-wise with the sentiment measures and the other explanatory
variables, \code{h} cannot be interpreted as the exact prediction horizon. 
In other words, \code{h} only shifts the input variables as they are
provided.} If \code{h = 0} (by default), no adjustments are made.  The
logical \code{do.difference} argument, if \code{TRUE}, can be used to
difference the target variable \code{y} supplied to the \code{sento_model()}
function, if it is a continuous variable (i.e.,~\code{model =
"gaussian"}).  The lag taken is the absolute value of the \code{h} argument
(given $\lvert h \rvert > 0$).  For example, if \code{h = 2}, and assuming
the \code{y} variable is aligned time-wise with all explanatory variables,
denoted by $X$ here for sake of the illustration, the regression performed
is of $y_{t+2} - y_{t}$ on $X_{t}$.  If $h = -2$, the regression fitted is
$y_{t+2} - y_{t}$ on $X_{t+2}$.
\item[$\bullet$] \code{do.iter}: To enact an iterative model estimation and
a one-step ahead out-of-sample analysis, set \code{do.iter = TRUE}.  To
perform a one-off in-sample estimation, set \code{do.iter = FALSE}.  The
arguments \code{nSample}, \code{start} and \code{nCore} are used for
iterative modeling, thus, when \code{do.iter = TRUE}.  The first argument
is $M$, that is, the size of the sample to re-estimate the model with each
time.  The second argument can be used to only run a later subset of the
iterations (\code{start = 1} by default runs all iterations).  The total
number of iterations is equal to \code{length(y)} - \code{nSample} -
\code{abs(h)} - \code{oos}, with \code{y} the response variable as a vector. 
The \code{oos} argument specifies partly, as explained above, the
cross-validation, but also provides flexibility in defining the
out-of-sample exercise.  For instance, given $t$, the one-step ahead
out-of-sample prediction is computed at $t + $\code{oos}$ + 1$.  As per
default, \code{oos = 0}.  If \code{nCore > 1}, the \code{\%dopar\%}
construct from the \proglang{R}~package \pkg{foreach} \citep{foreach} is
utilized to speed up the out-of-sample analysis.
\end{itemize}
To enhance the intuition about attribution, we estimate a contemporaneous
in-sample model and compute the attribution decompositions.
\begin{CodeChunk}
\begin{CodeInput}
R> ctrInSample <- ctr_model(model = "gaussian",
+    h = 0, type = "BIC", alphas = 0, do.iter = FALSE)
R> fit <- sento_model(sentMeas, y, ctr = ctrInSample)
\end{CodeInput}
\end{CodeChunk}
The \code{attributions()} function takes the `\code{sento_model}'
object and the related sentiment measures object as inputs, and generates by
default attributions for all in-sample dates at the level of individual
documents, lags, lexicons, features, and time weighting schemes.  The
function can be applied to a `\code{sento_modelIter}' object as well,
for any specific dates using the \code{refDates} argument.  If
\code{do.normalize = TRUE}, the values are normalized between $-1$ and $1$
through division by the $\ell_{2}$-norm of the attributions at a given
date.  The output is an `\code{attributions}' object.
\begin{CodeChunk}
\begin{CodeInput}
R> attrFit <- attributions(fit, sentMeas)
R> head(attrFit[["features"]])
\end{CodeInput}
\begin{CodeOutput}
         date      wsj     wapo  economy noneconomy uncertainty  election
1: 1995-07-24 0.001421 0.004144 -0.01380    0.01675   -0.002531 -0.004931
2: 1995-07-31 0.001899 0.004747 -0.01334    0.01739   -0.002468 -0.004824
3: 1995-08-07 0.001625 0.004348 -0.01257    0.01636   -0.002409 -0.004723
4: 1995-08-14 0.001241 0.004860 -0.01296    0.01590   -0.002353 -0.004231
5: 1995-08-21 0.001523 0.004248 -0.01178    0.01552   -0.002509 -0.004148
6: 1995-08-28 0.001431 0.004372 -0.01236    0.01583   -0.002462 -0.004071
\end{CodeOutput}
\end{CodeChunk}
Attribution decomposes a prediction into the different sentiment components
along a given dimension, for example, lexicons.  The sum of the individual
sentiment attributions per date, the constant, and other non-sentiment
measures are thus equal to the prediction.  Indeed, the piece of code below
shows that the difference between the prediction and the summed attributions
plus the constant is equal to zero throughout.
\begin{CodeChunk}
\begin{CodeInput}
R> X <- as.matrix(as.data.table(sentMeas)[, -1])
R> yFit <- predict(fit, newx = X)
R> attrSum <- rowSums(attrFit[["lexicons"]][, -1]) + fit[["reg"]][["a0"]]
R> all.equal(as.numeric(yFit), attrSum)
\end{CodeInput}
\begin{CodeOutput}
[1] TRUE 
\end{CodeOutput}
\end{CodeChunk}

\section{Application to predicting the CBOE Volatility
Index}\label{s:application}

A noteworthy amount of finance research has pointed out the impact of
sentiment expressed through various corpora on stock returns and trading
volume, including \citet{heston17}, \citet{jegadeesh13}, \citet{tetlock08},
and \citet{antweiler04}.  \citet{caporin17} create lexicon-based news
measures to improve daily realized volatility forecasts.  \citet{manela17}
explicitly construct a news-based measure closely related to the CBOE
Volatility Index (VIX) and a good proxy for uncertainty.  A more widely used
proxy for uncertainty is the Economic Policy Uncertainty (EPU) index
\citep{baker16}.  This indicator is a normalized text-based index of the
number of news articles discussing economic policy uncertainty, from ten
large U.S.\ newspapers.  A relationship between political uncertainty and
market volatility is found by \citet{pastor13}.

The VIX measures the annualized option-implied volatility on the S\&P 500
stock market index over the next 30 days.  It is natural to expect that
media sentiment and political uncertainty partly explain the expected
volatility measured by the VIX.  In this section, we test this using the EPU
index and sentiment variables constructed from the \code{usnews} dataset. 
We analyze if our textual sentiment approach is more helpful than the EPU
index in an out-of-sample exercise of predicting the end-of-month VIX in
six months.  The prediction specifications we are interested in are
summarized as follows:
\begin{align*}\label{eq:reg}
\mathcal{M}_{s}: && VIX_{u + h} &= \delta + \bgamma VIX_{u} +
\bbeta^\top \bs_{u} + \epsilon_{u + h}, && \\
\mathcal{M}_{epu}: && VIX_{u + h} &= \delta + \bgamma VIX_{u} +
\beta EPU_{u - 1} + \epsilon_{u + h}, && \\
\mathcal{M}_{ar}: && VIX_{u + h} &= \delta + \bgamma VIX_{u} +
\epsilon_{u + h}. &&
\end{align*}
The target variable $VIX_{u}$ is the most recent available end-of-month
daily VIX value.  We run the predictive analysis for $h = 6$ months.  The
sentiment time series are in $\bs_{u}$ and define the sentiment-based model
($\mathcal{M}_{s}$).  As primary benchmark, we exchange the sentiment
variables for $EPU_{u - 1}$, the level of the U.S.\ economic policy
uncertainty index in month $u - 1$ we know is fully available by month $u$
($\mathcal{M}_{epu}$).  We also consider a simple autoregressive
specification ($\mathcal{M}_{ar}$).

\begin{figure}[t!]
\centering
\includegraphics[width=\textwidth]{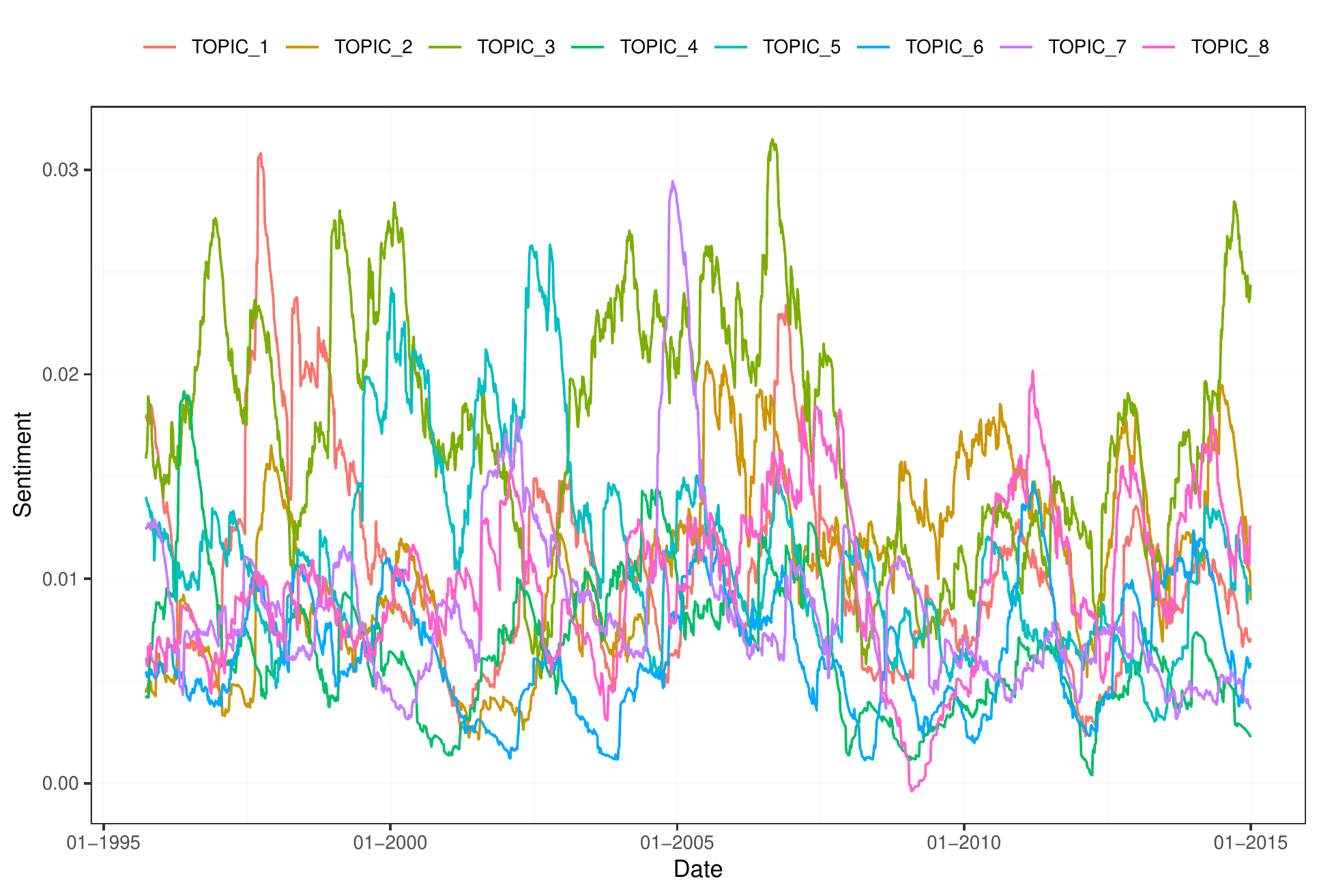}
\caption{Textual sentiment time series across latent topic features.}
\label{fig:sentmeasF}
\end{figure}

We use the built-in U.S.\ news corpus of around 4145 documents, in the
\code{uscorpus} object.  Likewise, we proceed with the nine lexicons and the
valence shifters list from the \code{lex} object used in previous examples. 
To infer textual features from scratch, we use a structural topic modeling
approach as implemented by the \proglang{R}~package \pkg{stm} \citep{stm}. 
This is a prime example of integrating a distinct text mining workflow with
our textual sentiment analysis workflow.  The \pkg{stm} package works with a
\pkg{quanteda} document-term matrix as an input.  We perform a fairly
standard cleaning of the document-term matrix, and use the default
parameters of the \code{stm()} function.  We group into eight features.
\begin{CodeChunk}
\begin{CodeInput}
R> dfm <- quanteda::tokens(uscorpus, what = "word", remove_punct = TRUE, 
+    remove_numbers = TRUE) %>%
+    quanteda::dfm(tolower = TRUE) %>%
+    quanteda::dfm_remove(quanteda::stopwords("en"), min_nchar = 3) %>%
+    quanteda::dfm_trim(min_termfreq = 0.95, termfreq_type = "quantile") %>%
+    quanteda::dfm_trim(max_docfreq = 0.15, docfreq_type = "prop") %>%
+    quanteda::dfm_subset(quanteda::ntoken(.) > 0)
R> topicModel <- stm::stm(dfm, K = 8, verbose = FALSE)
\end{CodeInput}
\end{CodeChunk}
We then define the keywords as the five most statistically representative
terms for each topic.  They are assembled in \code{keywords} as a list.
\begin{CodeChunk}
\begin{CodeInput}
R> topTerms <- t(stm::labelTopics(topicModel, n = 5)[["prob"]])
R> keywords <- lapply(1:ncol(topTerms), function(i) topTerms[, i])
R> names(keywords) <- paste0("TOPIC_", 1:length(keywords))
\end{CodeInput}
\end{CodeChunk}
We use the \code{add_features()} function to generate the features based on
the occurrences of these keywords in a document, scaling the feature values
between 0 and 1 by setting \code{do.binary = FALSE}.  We also delete all
current features.  Alternatively, one could use the predicted topics per
text as a feature and set \code{do.binary = TRUE}, to avoid documents
sharing mutual features, instead of relying on the generated keywords.  We
see a relatively even distribution of the corpus across the generated
features.
\begin{CodeChunk}
\begin{CodeInput}
R> uscorpus <- add_features(uscorpus, keywords = keywords, 
+    do.binary = FALSE, do.regex = FALSE)
R> quanteda::docvars(uscorpus, c("uncertainty", "election", 
+    "economy", "noneconomy", "wsj", "wapo")) <- NULL
R> colSums(quanteda::docvars(uscorpus)[, -1] != 0)
\end{CodeInput}
\begin{CodeOutput}
TOPIC_1 TOPIC_2 TOPIC_3 TOPIC_4 TOPIC_5 TOPIC_6 TOPIC_7 TOPIC_8 
   1111    1101    1389     648     954    1005    1052     856
\end{CodeOutput}
\end{CodeChunk}
The frequency of our target variable is monthly, yet we aggregate the
sentiment time series on a daily level and then across time using a lag of
about nine months (\code{lag = 270}).  While the lag is substantial, we also
include six different Beta time weighting schemes to capture time dynamics
other than a slowly moving trend.
\begin{CodeChunk}
\begin{CodeInput}
R> ctrAggPred <- ctr_agg(howWithin = "proportionalPol", 
+    howDocs = "equal_weight", howTime = "beta",  
+    by = "day", fill = "latest", lag = 270, aBeta = 1:3, bBeta = 1:2)
R> sentMeasPred <- sento_measures(uscorpus, lexicons = lex, ctr = ctrAggPred)
\end{CodeInput}
\end{CodeChunk}
In Figure~\ref{fig:sentmeasF}, we see sufficiently differing average time
series patterns for all topics.  The drop in sentiment during the recent
financial crisis is apparent, but the recovery varies along features.

\begin{figure}[t!]
\centering
\includegraphics[width=\textwidth]{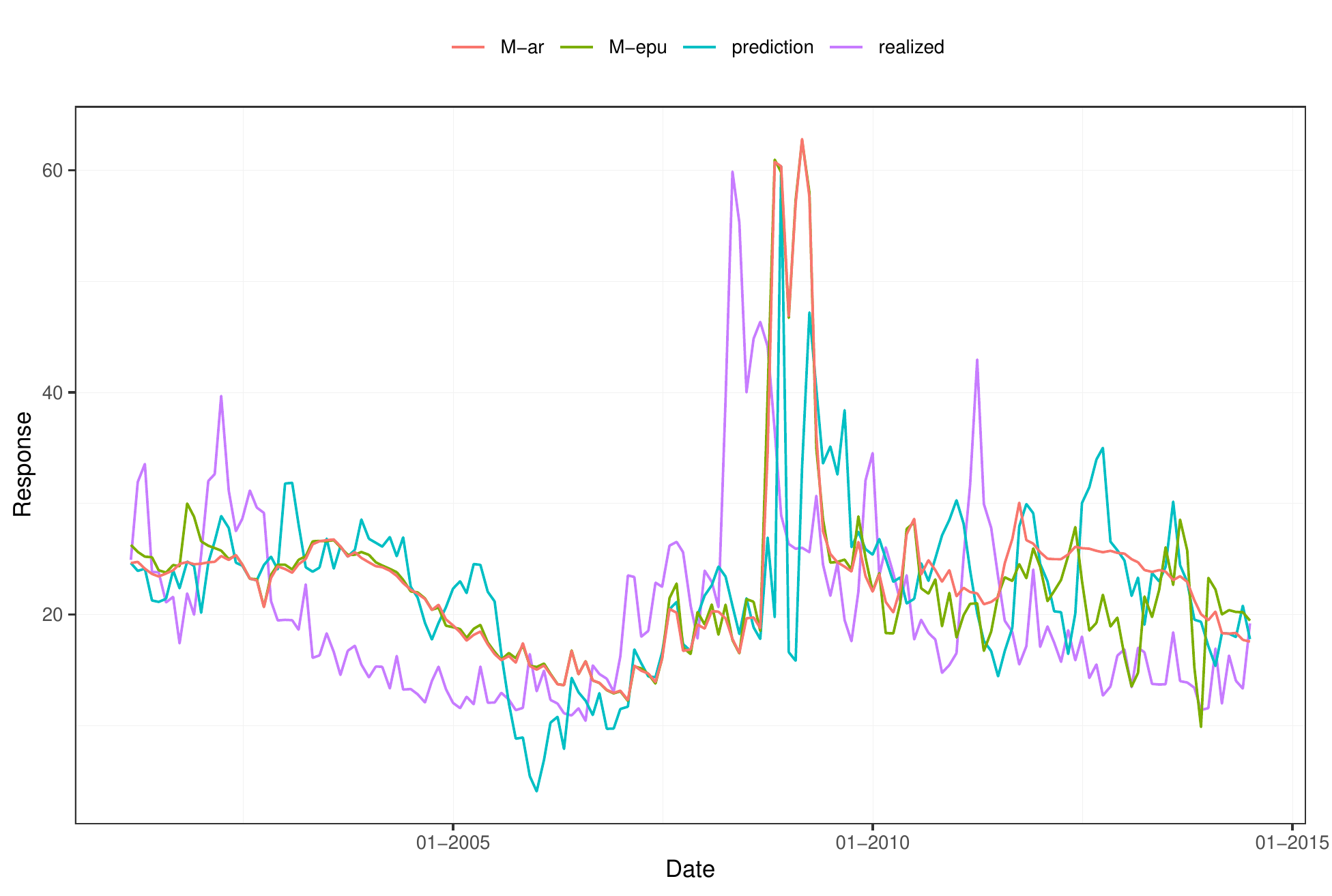}
\caption{Realized six-month ahead $VIX_{u + 6}$ index values (purple,
\code{realized}) and out-of-sample predictions from models $\mathcal{M}_{s}$ (blue, \code{prediction}), $\mathcal{M}_{ar}$ (red), and $\mathcal{M}_{epu}$ (green).}
\label{fig:for}
\end{figure}

The package has the EPU index as a dataset \code{epu} included.  We consider
the EPU index values one month before all other variables and use the
\pkg{lubridate} package \citep{lubridate} to do this operation.  We assure
that the length of the dependent variable is equal to the number of
observations in the sentiment measures by selecting based on the proper
monthly dates.  The pre-processed VIX data is represented by the \code{vix}
variable.\footnote{The VIX data is retrieved from
\url{https://fred.stlouisfed.org/series/VIXCLS}.}
\begin{CodeChunk}
\begin{CodeInput}
R> data("epu", package = "sentometrics")
R> sentMeasIn <- subset(sentMeasPred, date %in% vix$date)
R> datesIn <- get_dates(sentMeasIn)
R> datesEPU <- lubridate::floor_date(datesIn, "month") %m-% months(1)
R> xEPU <- epu[epu$date %in% datesEPU, "index"]
R> y <- vix[vix$date %in% datesIn, "value"]
R> x <- data.frame(lag = y, epu = xEPU)
\end{CodeInput}
\end{CodeChunk}
We apply the iterative rolling forward analysis (\code{do.iter = TRUE}) for
our 6-month prediction horizon.  The target variable is aligned with the
sentiment measures in \code{sentMeasIn}, such that \code{h = 6} in the
modeling control means forecasting the monthly averaged VIX value in six
months.  The calibration of the sparse linear regression is based on a
Bayesian-like information criterion (\code{type = "BIC"}) proposed by
\citet{tibshirani12}.  We configure a sample size of $M = 60$ months for a
total sample of $N = 232$ observations.  Our out-of-sample setup is
nonoverlapping; \code{oos = h - 1} means that for an in-sample estimation
at time $u$, the last available explanatory variable dates from time $u -
h$, and the out-of-sample prediction is performed at time $u$ as well, not
at time $u - h + 1$.  We consider a range of \code{alpha} values that allows
any of the Ridge, LASSO, and pure elastic net regularization objective
functions.\footnote{When a sentiment measure is a duplicate of another or
when at least 50\% of the series observations are equal to zero, it is
automatically discarded from the analysis.  Discarded measures are put under
the \code{"discarded"} element of a `\code{sento\_model}' object.}
\begin{CodeChunk}
\begin{CodeInput}
R> h <- 6
R> oos <- h - 1
R> M <- 60
R> ctrIter <- ctr_model(model = "gaussian",
+    type = "BIC", h = h, alphas = c(0, 0.1, 0.3, 0.5, 0.7, 0.9, 1),
+    do.iter = TRUE, oos = oos, nSample = M, nCore = 1)
R> out <- sento_model(sentMeasIn, x = x[, "lag", drop = FALSE], y = y, 
+    ctr = ctrIter)
R> summary(out)
\end{CodeInput}
\begin{CodeOutput}
Model specification 
- - - - - - - - - - - - - - - - - - - - 
 
Model type: gaussian 
Calibration: via BIC information criterion 
Sample size: 60 
Total number of iterations/predictions: 161 
Optimal average elastic net alpha parameter: 0.92 
Optimal average elastic net lambda parameter: 2.8 
 
Out-of-sample performance 
- - - - - - - - - - - - - - - - - - - - 
 
Mean directional accuracy: 47.5 % 
Root mean squared prediction error: 10.42 
Mean absolute deviation: 8.2
\end{CodeOutput}
\end{CodeChunk}
The output of the \code{sento_model()} call is a
`\code{sento_modelIter}' object.  Below we replicate the analysis for
the benchmark regressions, without the sentiment variables.  The vector
\code{preds} assembles the out-of-sample predictions for model
$\mathcal{M}_{epu}$, the vector \code{predsAR} for model $\mathcal{M}_{ar}$.
\begin{CodeChunk}
\begin{CodeInput}
R> preds <- predsAR <- rep(NA, nrow(out[["performance"]]$raw))
R> yTarget <- y[-(1:h)]
R> xx <- x[-tail(1:nrow(x), h), ]
R> for (i in 1:(length(preds))) {
+    j <- i + M    
+    data <- data.frame(y = yTarget[i:(j - 1)], xx[i:(j - 1), ])
+    reg <- lm(y ~ ., data = data)
+    preds[i] <- predict(reg, xx[j + oos, ])
+    regAR <- lm(y ~ ., data = data[, c("y", "lag")])
+    predsAR[i] <- predict(regAR, xx[j + oos, "lag", drop = FALSE])
R> }
\end{CodeInput}
\end{CodeChunk}
A more detailed view of the different performance measures, in this case
directional accuracy, root mean squared, and absolute errors, is obtained
via \code{out[["performance"]]}.  A list of the individual
`\code{sento_model}' objects is found under \code{out[["models"]]}. 
A simple plot to visualize the out-of-sample fit of any
`\code{sento_modelIter}' object can be produced using \code{plot()}. 
We display in Figure~\ref{fig:for} the realized values and the different
predictions.
\begin{CodeChunk}
\begin{CodeInput}
R> plot(out) +
+    geom_line(data = melt(data.table(date = names(out$models), "M-epu" = 
+    preds, "M-ar" = predsAR, check.names = FALSE), id.vars = "date"))
\end{CodeInput}
\end{CodeChunk}
Table~\ref{table:performance} reports two common out-of-sample prediction
performance measures, decomposed in a pre-crisis period (spanning up to
June 2007, from the point of view of the prediction date), a crisis period
(spanning up to December 2009) and a post-crisis period.  It appears that
sentiment adds predictive power during the crisis period.  The flexibility
of the elastic net avoids that predictive power is too seriously compromised
when adding sentiment to the regression, even when it has no added value.

\begin{figure}[t!]
\centering
\includegraphics[width=\textwidth]{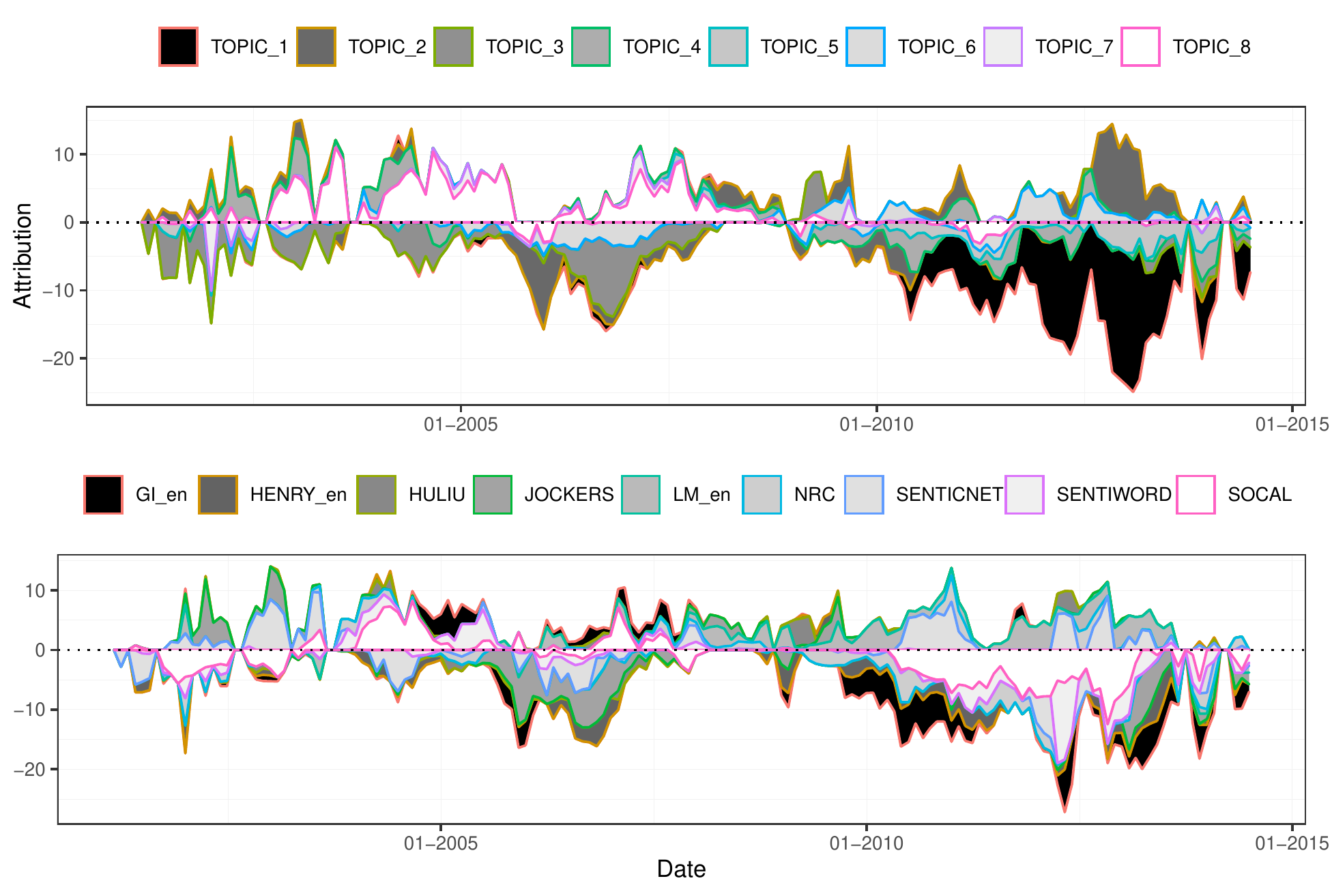}
\caption{Attribution to features (top) and lexicons (below).}
\label{fig:attr}
\end{figure}

\begin{table}[t!]
\centering
\begin{tabular}{lccccccccccccccc}
\toprule[1pt]
& \multicolumn{3}{c}{Full} && \multicolumn{3}{c}{Pre-crisis} &&
\multicolumn{3}{c}{Crisis} && \multicolumn{3}{c}{Post-crisis} \\
\cmidrule{2-4} \cmidrule{6-8} \cmidrule{10-12} \cmidrule{14-16}   
& $\mathcal{M}_{s}$ & $\mathcal{M}_{epu}$ & $\mathcal{M}_{ar}$ && $\mathcal{M}_{s}$ & $\mathcal{M}_{epu}$ & $\mathcal{M}_{ar}$ && $\mathcal{M}_{s}$ & $\mathcal{M}_{epu}$ & $\mathcal{M}_{ar}$ && $\mathcal{M}_{s}$ & $\mathcal{M}_{epu}$ & $\mathcal{M}_{ar}$ \\
\toprule[1pt]
\multirow{1}{*}{RMSE} & 10.42 & 10.42 & 10.72 & & 7.55 & 6.64 & 6.52 & & 16.73 &
19.23 & 19.46 & & 9.29 & 7.43 & 8.42 \\
\multirow{1}{*}{MAD} & 8.20 & 7.46 & 7.93 & & 6.58 & 5.78 & 5.60 & & 13.20 &
14.23 & 14.56 & & 7.72 & 6.09 & 7.54 \\
\bottomrule[1pt]      
\end{tabular}
\caption{Performance measures.  The root mean squared error (RMSE) is
computed as $\sum_{i = 1}^{N_{oos}}e_{i}^2 / N_{oos}$, and the mean absolute
deviation (MAD) as $\sum_{i = 1}^{N_{oos}}\lvert e_{i} \rvert / N_{oos}$,
with $N_{oos}$ the number of out-of-sample predictions, and $e_{i}$ the
prediction error.  The sentiment-based model is $\mathcal{M}_{s}$, the
EPU-based model is $\mathcal{M}_{epu}$, the autoregressive model is
$\mathcal{M}_{ar}$.}
\label{table:performance}
\end{table}
The last step is to perform a post-modeling attribution analysis.  For a
`\code{sento_modelIter}' object, the \code{attributions()} function
generates sentiment attributions for all out-of-sample dates.  To study
the evolution of the prediction attribution, the attributions can be
visualized with the \code{plot()} function applied to the
`\code{attributions}' output object.  This can be done according to
any of the dimensions, except for individual documents. 
Figure~\ref{fig:attr} shows two types of attributions in separate panels. 
The attributions are displayed stacked on top of each other, per date.  The
$y$-axis represents the attribution to the prediction of the target variable. 
The third topic was most impactful during the crisis, and the first topic
received the most negative post-crisis weight.  Likewise, the lexicons
attribution conveys an increasing influence of the SO-CAL lexicon on the
predictions.  Finally, it can be concluded that the predictive role of
sentiment is least present before the crisis.

This illustration shows that the \pkg{sentometrics} package provides useful
insights in predicting variables like the VIX starting from a corpus of
texts.  Results could be improved by expanding the corpus, or by optimizing
the features generation.  For a larger application of the entire workflow,
we refer to \citet{ardia19}.  They find that the incorporation of textual
sentiment indices results in better prediction of the U.S.\ industrial
production growth rate compared to using a panel of typical macroeconomic
indices only.

\section{Conclusion and future development}\label{s:conclusion}

The \proglang{R}~package \pkg{sentometrics} provides a framework to
calculate sentiment for texts, to aggregate textual sentiment scores into
many time series at a desired frequency, and to use these in a flexible
prediction modeling setup.  It can be deployed to quantify a qualitative
corpus of texts, relate it to a target variable, and retrieve which type of
sentiment is most informative through visualization and attribution
analysis.  The main priorities for further development are integrating
better prediction tools, enhancing the complexity of the sentiment engine,
allowing user-defined weighting schemes, and adding intra-day aggregation.

If you use \proglang{R} or \pkg{sentometrics}, please cite the software in
publications.  In case of the latter, use \code{citation("sentometrics")}. 
Additional code examples can be found in the regularly updated ``Examples''
section at \url{https://sborms.github.io/sentometrics}.

\section*{Computational details}

The results in this paper were obtained using \proglang{R} 4.1.0 \citep{R},
\pkg{sentometrics} version 0.8.4, and underlying or used packages
\pkg{caret} version 6.0.88 \citep{caret}, \pkg{data.table} version 1.14.0
\citep{dt}, \pkg{foreach} version 1.5.1 \citep{foreach}, \pkg{ggplot2}
version 3.3.4 \citep{ggplot2}, \pkg{glmnet} version 4.1.1 \citep{glmnet},
\pkg{gridExtra} version 2.3.0 \citep{gridextra}, \pkg{ISOweek} version 0.6.2
\citep{isoweek}, \pkg{lexicon} version 1.2.1 \citep{lexicon},
\pkg{lubridate} version 1.7.10 \citep{lubridate}, \pkg{quanteda} version
3.0.0 \citep{quanteda}, \pkg{Rcpp} version 1.0.6 \citep{rcpp},
\pkg{RcppArmadillo} version 0.10.5.0.0 \citep{rcpparmadillo},
\pkg{RcppRoll} version 0.3.0 \citep{rcpproll}, \pkg{RcppParallel} version
5.1.4 \citep{rcppparallel}, \pkg{stm} version 1.3.6 \citep{stm},
\pkg{stringi} version 1.6.2 \citep{stringi}, \pkg{tm} version 0.7.8
\citep{tm}, and \pkg{zoo} version 1.8.9 \citep{zoo}.  Computations were
performed on a Windows 10 Pro machine, x86 64--w64--mingw32/x64 (64--bit)
with Intel(R) Core(TM) i7--7700HQ CPU 2x 2.80 GHz.  The code used in the
main paper is available in the \proglang{R} script \code{run\_vignette.R}
located in the \code{examples} folder on the dedicated \pkg{sentometrics}
\proglang{GitHub} repository at \url{https://github.com/SentometricsResearch/sentometrics}.  \proglang{R}, \pkg{sentometrics}, and all other packages are available from
the Comprehensive \proglang{R} Archive Network (CRAN) at
\url{https://CRAN.R-project.org}.  Any version under development will be available on our \proglang{GitHub} repository. Additional resources related to \emph{sentometrics} can be found at \url{https://sentometrics-research.com}.

\section*{Acknowledgments}

We thank the Associate Editors (Toby Hocking and Torsten Hothorn) and three
anonymous Referees, Andres Algaba (\emph{package contributor}), Nabil
Bouamara, Peter Carl, Leopoldo Catania, Thomas Chuffart, Dries Cornilly,
Serge Darolles, William Doehler, Arnaud Dufays, Matteo Ghilotti, Kurt
Hornik, Siem Jan Koopman, Julie Marquis, Linda Mhalla, Brian Peterson, Laura
Rossetti, Tobias Setz, Majeed Siman, Stefan Theussl, Wouter Torsin, Jeroen
Van Pelt (\emph{package contributor}), Marieke Vantomme, and participants
at the CFE (London, 2017), eRum (Budapest, 2018), \proglang{R}/Finance (Chicago, 2018),
SwissText (Winterthur, 2018), SoFiE (Brussels, 2018), ``Data Science in
Finance with \proglang{R}'' (Vienna, 2018), ``New Challenges for Central Bank
Communication'' (Brussels, 2018), (EC)\^{}2 (Roma, 2018), and use\proglang{R}! 
(Toulouse, 2019) conferences for helpful comments.  We acknowledge Google
Summer of Code 2017 and 2019 (\url{https://summerofcode.withgoogle.com}),
Innoviris (\url{https://innoviris.brussels}), IVADO
(\url{https://ivado.ca}), and the Swiss National Science
Foundation (\url{http://www.snf.ch}, grants \#179281 and \#191730) for their
financial support.

\bibliography{ref}

\newpage
\begin{appendix}

\section[Package methods overview]{Package methods
overview}\label{appendix:methods}

This appendix provides an overview of the \proglang{R} methods made
available in the \pkg{sentometrics} package, as also highlighted in
Table~\ref{table:taxonomy}.  The \proglang{S}3 class objects from the
\pkg{sentometrics} package are created using their function counterpart with
the same name, except for the `\code{sentiment}' object (created with
the \code{compute_sentiment()} function) and the
`\code{sento_modelIter}' object (created with the
\code{sento_model(\dots, ctr = ctr_model(\dots, do.iter = TRUE))} function). 
Most of the methods are individually documented; to access the help files do
\code{?method.object} (e.g.,~\code{?aggregate.sento_measures}).
\begin{table}[h!]
\centering
\begin{tabular}{rL{11.6cm}}
\multicolumn{1}{l}{\textbf{Standard methods}} & \\
\code{plot()} & Classes: `\code{attributions}', `\code{sento_measures}',
`\code{sento_modelIter}'. \newline
Plots, all in a similar \pkg{ggplot2} style, respectively, the computed
sentiment attributions of a run regression model, the constructed sentiment
measures, and the target variable versus the predicted outcomes of an
iteratively ran regression model. The first two can be grouped according a
specific dimension (e.g.,~by \code{"features"}). \\
\code{summary()} & Classes: `\code{sento_measures}, `\code{sento_model}',
`\code{sento_modelIter}'. \newline
Provides a short description of the contents of the respective object.
The \code{print()} method simply displays the object class; it is also
supported for a `\code{sento_corpus}' object and prints like in
\pkg{quanteda}. \\
& \\
\multicolumn{1}{l}{\textbf{Statistical methods}} & \\
\code{aggregate()} & Classes: `\code{sentiment}', `\code{sento_measures}. \newline
Aggregates a document-level or a sentence-level `\code{sentiment}'
object into a `\code{sento_measures}' object, or a sentence-level
`\code{sentiment}' object into a document-level
`\code{sentiment}' object.  \newline
A `\code{sento_measures}' object
can also be further aggregated across sentiment measures.  \\
\code{diff()} & Classes: `\code{sento_measures}'. \newline
Returns a `\code{sento_measures}' object with differenced sentiment
measures.  \\
\code{merge()} & Classes: `\code{sentiment}'. \newline
Combines multiple `\code{sentiment}' objects row-wise and/or
column-wise.  \\
\code{nobs()} & Classes: `\code{sento_measures}'. \newline
Gives the number of data points (i.e.,~rows) in the sentiment
measures.  The number of sentiment measures can be obtained with the
\code{nmeasures()} function.  \\
\code{predict()} & Classes: `\code{sento_model}'. \newline
Generates predictions from the model object for a data
`\code{matrix}' of values for the explanatory sentiment measures and
other variables.  \\
\code{scale()} & Classes: `\code{sento_measures}'. \newline
Returns a `\code{sento_measures}' object with scaled sentiment
measures.  One can also use the \code{center} and \code{scale} arguments to
define values to add to the sentiment measures or divide them by.
\end{tabular}
\end{table}
\begin{table}[t!]
\centering
\begin{tabular}{rL{11.6cm}}
\multicolumn{1}{l}{\textbf{Coercion}} & \\
\code{as.data.frame()} & Classes: `\code{sento_corpus}', `\code{sento_measures}'. \newline
Converts the corpus or sentiment measures in a `\code{data.frame}' object. \\
\code{as.data.table()} & Classes: `\code{sento_corpus}', `\code{sento_measures}'. \newline
Converts the corpus or sentiment measures in a `\code{data.table}' object. \\
\code{as.sentiment()} & Classes: `\code{data.frame}', `\code{data.table}'. \newline
Converts a properly structured sentiment table in `\code{data.frame}' or
`\code{data.table}' format into a document-level or a sentence-level
`\code{sentiment}' object. \\
\code{as.sento_corpus()} & Classes: \pkg{quanteda} `\code{corpus}', \pkg{tm}
`\code{SimpleCorpus}', \pkg{tm} `\code{VCorpus}'. \newline
Transforms the given corpus input object into a `\code{sento_corpus}'
object, integrating available metadata, where possible, into corpus features. \\
& \\
\multicolumn{1}{l}{\textbf{Extraction}} & \\
\code{subset()} & Classes: `\code{sento_measures}'. \newline
Can be used to do three things: subset the rows (either by index or by a
condition), select certain sentiment measures, or delete certain sentiment
measures.  The selection and deletion is based on the names of the sentiment
measures along the features, lexicons, and time-weighting schemes
dimensions.
\end{tabular}
\end{table}

\section[Aggregation weighting schemes]{Aggregation weighting
schemes}\label{appendix:weighting}

This appendix presents the formulas that define the weights used in the
different sentiment aggregation schemes available in the package.  The
constant $c$ indicates a normalization factor that makes sure the considered
weights sum up to 1.  When not specified, arguments referred to are from the
\code{ctr_agg()} function.

\subsection*{Within-document and within-sentence weighting}

We outline here the different options available for the \code{howWithin}
argument of the \code{ctr_agg()} function and the \code{how} argument of the
\code{compute_sentiment()} function, for the sentiment calculation in
\eqref{eq:aggwithin}.  The weight $\omega_{i}$ is associated to the unigram
at the $i$th position in a document (resp.\ sentence) $d_{n, t}$, where $d$
serves as a notational shorthand.  The number of unigrams in a document
(resp.\ sentence) is $Q_{d}$, the number of unigrams in a document (resp.\
sentence) that appear in the lexicon is $n_{pol}$, $N$ is the total number
of documents (resp.\ sentences) in the corpus, and $q_{i}$ is the number of
documents (resp.\ sentences) across the entire corpus containing unigram
$i$.

The package lets the user choose between the following constant and
unigram-specific weights:
\begin{itemize}
\item \code{"counts"}: 
\begin{equation*}
\omega_{i} = 1
\end{equation*}
\item \code{"proportional"}: 
\begin{equation*}
\omega_{i} = \frac{1}{Q_{d}}
\end{equation*}
\item \code{"proportionalPol"}: 
\begin{equation*}
\omega_{i} = \frac{1}{\max\{n_{pol}, 1\}}
\end{equation*}
\item \code{"proportionalSquareRoot"}: 
\begin{equation*}
\omega_{i} = \frac{1}{\sqrt{Q_{d}}}
\end{equation*}
\item \code{"UShaped"}: 
\begin{equation*}
\omega_{i} = \left( i - \frac{Q_{d} + 1}{2} \right) ^2 \times c
\end{equation*}
\item \code{"inverseUShaped"}: 
\begin{equation*}
\omega_{i} = \left( 0.25 - \frac{\left( i - \frac{Q_{d} + 1}{2} \right) ^2}{Q_{d}^2} \right) \times c
\end{equation*}
\item \code{"exponential"}: 
\begin{equation*}
\omega_{i} = \exp \left( 5 \left( \frac{i}{Q_{d}} - 1 \right) \right) \times c
\end{equation*}
\item \code{"inverseExponential"}: 
\begin{equation*}
\omega_{i} = \exp \left( 5 \left( 1 - \frac{i}{Q_{d}} \right) \right) \times c
\end{equation*}
\item \code{"TFIDF"}: 
\begin{equation*}
\omega_{i} = \log_{10} \left( \frac{N}{1 + q_{i}} \right)
\end{equation*}
\end{itemize}
\emph{Note:} The \code{"TFIDF"} option represents term frequency-inverse
document frequency weighting \citep{Sparck1972}.  The weight covers only the
inverse document frequency (IDF) part, and we follow the implementation of
the \pkg{quanteda} package's \code{docfreq(\dots, scheme = "inverse", k = 1,
base = 10)} function.  The term frequency (TF) component is inherent in \eqref{eq:aggwithin}; for instance, it will pertain to the raw
count convention when using no valence shifters.

\subsection*{Across-document and across-sentence weighting}

We outline here the different options available for the \code{howDocs}
argument of the \code{ctr_agg()} function, for the aggregation in
\eqref{eq:aggdocuments}.  The weight $\theta_{n}$ values a document (resp.\
sentence) $d_{n, t}$ (again, we use $d$ as a shorthand) in the aggregation
window (per date for across-document, and per document for
across-sentence).  Recall that $N_{t}$ is the total number of documents at
time $t$, or, abusing the notation, it can similarly represent the number of
sentences within a document.  The total number of unigrams of all documents
(resp.\ sentences) included in the aggregation window is $z$.

For a given document (resp.\ sentence) $d_{n, t}$, the weight can be one of:
\begin{itemize}
\item \code{"equal_weight"}: 
\begin{equation*}
\theta_{n} = \frac{1}{N_{t}}
\end{equation*}
\item \code{"proportional"}: 
\begin{equation*}
\theta_{n} = Q_{d} \times c
\end{equation*}
\item \code{"inverseProportional"}: 
\begin{equation*}
\theta_{n} = \frac{1}{Q_{d}} \times c
\end{equation*}
\item \code{"exponential"}: 
\begin{equation*}
\theta_{n} = \exp \left( \alpha \left( \frac{Q_{d}}{z} - 1 \right) \right) \times c
\end{equation*}
\item \code{"inverseExponential"}: 
\begin{equation*}
\theta_{n} = \exp \left( \alpha \left( 1 - \frac{Q_{d}}{z} \right) \right) \times c
\end{equation*}
\end{itemize}

The value $\alpha$ is set equal to $10$ $\times$ \code{alphaExpDocs}.

\subsection*{Across-time weighting}

We outline here the options available for the \code{howTime} argument of the
\code{ctr_agg()} function, for the aggregation in \eqref{eq:aggtime}.  The
weight $b_t$ represents the time weight for a sentiment value at the time
point $t$ relative to a starting date and a lag $\tau$.

For positional time points $t = 1, \ldots, \tau$, the weighting schemes
available are:
\begin{itemize}
\item \code{"equal_weight"}: 
\begin{equation*}
b_{t} = \frac{1}{\tau}
\end{equation*}
\item \code{"almon"}: 
\begin{equation*}
b_{t} = \left[ \left( 1 - \frac{t}{\tau} \right) ^{R-r}
\left( 1 -  \left( 1 - \frac{t}{\tau} \right) ^r \right) \right] \times c
\end{equation*}
\item \code{"beta"}: 
\begin{equation*}
b_{t} = f  \left( \frac{t}{\tau}; a, b \right) / \sum_{t=1}^{\tau}f
\left( \frac{t}{\tau}, a, b \right) \times c, \:\text{where}\: 
\end{equation*}
\begin{equation*}
\setlength{\abovedisplayskip}{0pt}
\setlength{\abovedisplayshortskip}{0pt}
f(x; a, b) \bydef \frac{x^{a-1}(1-x)^{b-1}\Gamma(a+b)}{\Gamma(a)\Gamma(b)},
\:\text{and}\: \Gamma(\cdot) \:\text{is the gamma function}\:
\end{equation*}
\item \code{"linear"}: 
\begin{equation*}
b_{t} = \frac{t}{\tau} \times c
\end{equation*}
\item \code{"exponential"}: 
\begin{equation*}
b_{t} = \exp \left( \alpha \left( \frac{t}{\tau} - 1 \right) \right) \times c
\end{equation*}
\end{itemize}

The value $r$ is a specific element from the \code{ordersAlm} vector, and
$R$ is the maximum value in that vector.  If \code{do.inverseAlm = TRUE},
the inverse Almon polynomials are computed too, modifying $\frac{t}{\tau}$
to $1-\frac{t}{\tau}$.  If \code{do.inverseExp = TRUE}, the inverse
exponential curves are added.  In the Beta density $f(\cdot; \cdot, \cdot)$,
$a$ is \code{aBeta}, and $b$ is \code{bBeta}.  The $\alpha$ here is defined
as $10$ $\times$ \code{alphasExp}.  The functions \code{weights\_almon()},
\code{weights\_beta()} and \code{weights\_exponential()} allow generating
these time weights separately.
\end{appendix}

\end{document}